%% file: main.tex
\documentclass[10pt,twocolumn,letterpaper]{article}

\usepackage{iccv}
\usepackage{times}
\usepackage{epsfig}
\usepackage{graphicx}
\usepackage{amsmath}
\usepackage{amssymb}
\usepackage{booktabs}
\usepackage{multirow}
\usepackage{makecell}
\usepackage[linesnumbered,ruled,vlined]{algorithm2e}
\usepackage{algpseudocode}
\usepackage{comment}
\SetKwInput{KwInput}{Input} 
\usepackage[table]{xcolor}

\usepackage[pagebackref=true,breaklinks=true,letterpaper=true,colorlinks,bookmarks=false]{hyperref}
\usepackage{amsmath}
\usepackage{amssymb}
\usepackage{booktabs}
\usepackage{multirow}
\usepackage{makecell}
\usepackage[table]{xcolor}
\usepackage{pifont}%
\newcommand{\cmark}{\ding{51}}%
\newcommand{\xmark}{\ding{55}}%
\usepackage{subcaption}
\usepackage[accsupp]{axessibility}

\usepackage[capitalize]{cleveref}
\crefname{section}{Sec.}{Secs.}
\Crefname{section}{Section}{Sections}
\Crefname{table}{Table}{Tables}
\crefname{table}{Tab.}{Tabs.}

\definecolor{codeblue}{rgb}{0.25,0.5,0.5}

\SetCommentSty{mycommfont}
\newcommand{\condenseparagraph}[1]{\vspace{.20em}\noindent\textbf{#1}~}

\input{macros.tex}

\iccvfinalcopy %

\newcommand{\mb}[1]{\ensuremath{\mathbf{#1}}}
\newcommand{\mc}[1]{\ensuremath{\mathcal{#1}}}

\begin{document}

\title{SceneRF: Self-Supervised Monocular 3D Scene Reconstruction \\with Radiance Fields}
\author{
    Anh-Quan Cao, Raoul de Charette \\
	Inria \\
	\url{https://astra-vision.github.io/SceneRF}
}
\maketitle
\ificcvfinal\thispagestyle{empty}\fi

\begin{abstract}
3D reconstruction from a single 2D image was extensively covered in the literature but relies on depth supervision at training time, which limits its applicability.
To relax the dependence to depth
we propose SceneRF, a self-supervised monocular scene reconstruction method using only posed image sequences for training. 
Fueled by the recent progress in neural radiance fields (NeRF) we optimize a radiance field though with explicit depth optimization and a novel probabilistic sampling strategy to efficiently handle large scenes. 
At inference, a single input image suffices to hallucinate novel depth views which are fused together to obtain 3D scene reconstruction.
Thorough experiments demonstrate that we outperform all baselines for novel depth views synthesis and scene reconstruction, on indoor BundleFusion and outdoor SemanticKITTI. Code is available at \emph{\href{https://astra-vision.github.io/SceneRF}{https://astra-vision.github.io/SceneRF}}~.
\end{abstract}

\section{Introduction}
\label{sec:intro}

Humans evolve in a 3D physical world where even the slightest motion requires a thorough understanding of their surroundings to avoid collisions. While binocular vision is an evident evolutionary edge, physiological studies suggest that humans can sense depth even with monocular vision~\cite{koenderink1995depth}. Despite a long-standing line of research~\cite{VANDENHEUVEL1998354, 790426, sturm1999method} this is yet unequaled by computer vision algorithms, which mostly rely on multiple-views to reconstruct complex scenes~\cite{roessle2022depthpriorsnerf}. However, estimating 3D from a single view would unveil novel applications in a world flooded with consumer cameras where mobile robots, like autonomous cars, still require costly depth sensors~\cite{Caesar2020nuScenesAM,semkitti}.

A small portion of the 3D field addressed reconstruction of complex scenes from a single image~\cite{huang2018holistic, zhang2021holistic, monoscene, dahnert2021panoptic} but they all require depth supervision which discourage acquisition of image-only datasets.
Meanwhile, Neural Radiance Field~\cite{nerf} (NeRF), which optimizes a radiance field self-supervisedly from one or more views, unraveled many descendants~\cite{xie2022neural} with unprecedented performance on novel views  synthesis. They are however, mostly limited to objects when it comes to single-view input~\cite{Liu20222DGM, Niemeyer2020DifferentiableVR,Mller2022AutoRFL3}. 
For complex scenes, besides~\cite{mine2021} all train on synthetic data~\cite{sharma2022seeing} or require additional geometrical cues to train on real data~\cite{urbannerf,dsnerf,roessle2022depthpriorsnerf}.
Reducing the need of supervision on complex scenes would lower our dependency to costly-acquired datasets.

\begin{figure}
	\centering
	\footnotesize
	\includegraphics[width=0.95\linewidth]{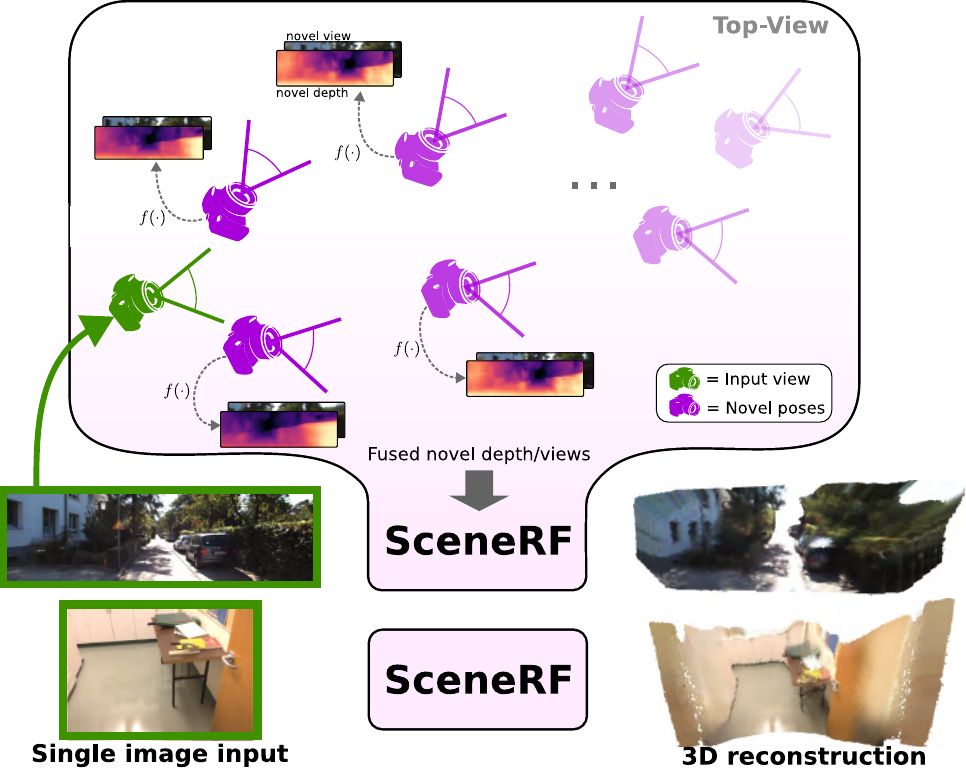}
	\caption{\textbf{SceneRF overview.} 
	From a single {\bf\color{inputcol}input image}, SceneRF synthesizes {\bf\color{novelcol}novel depth/views}, at arbitrary poses, which are then fused to estimate 3D reconstruction. It relies on an image-conditioned NeRF (here, $f(\cdot)$) trained self-supervisedly on image sequences with pose.}
	\label{fig:teaser}
\end{figure}

In this work, we address single-view reconstruction of complex (and possibly large) scenes, in a fully self-supervised manner. 
SceneRF trains only with sequences of posed images to optimize a \textit{large} neural radiance fields (NeRF). \cref{fig:teaser} illustrates inference where a single RGB image suffices to reconstruct the 3D scene from the fusion of synthesized novel depths/views, sampled at arbitrary locations.
We build upon PixelNeRF~\cite{pixelnerf} and propose specific design choices to \textit{explicitly} optimize depth. 
Because large scenes hold their own challenges, we introduce a novel probabilistic ray sampling to efficiently choose the sparse locations to optimize within the large radiance volume, and introduce a Spherical U-Net, which aims is to enable hallucination beyond the input image field of view. We summarize our contributions below:
\begin{itemize}
	\item We build on custom design choices to explicitly optimize depth (\cref{sec:nerf_for_depth}) with a Spherical U-Net (\cref{sec:sphere}) -- altogether allowing use of our radiance field for scene reconstruction (\cref{sec:reconstruction}),
	\item Our probabilistic ray sampling (\cref{sec:ray_sampling}) learns to model the continuous density volume with a mixture of Gaussians --~boosting both performance and efficiency,
	\item To the best of our knowledge, we propose the first self-supervised large scene reconstruction method using a single-view as input. Results on indoor and driving scenes show that SceneRF even outperforms depth-supervised baselines (\cref{sec:experiments}).
\end{itemize}

\section{Related work}
\label{sec:related_work}

As the 3D literature recently blossomed with the rise of NeRF methods~\cite{xie2022neural}, 
we limit our review to the smaller portion of works using a \textbf{\makebox{single} input view}, and study the literature along two axes related to our work: \textit{novel views/depths synthesis} and \textit{3D~reconstruction}. %
\paragraph*{Novel views/depths synthesis.} 
Rendering novel views from an image has been a long-lasting research problem~\cite{horry1997tour,tatarchenko2015single,kenburn3d,xu_disn_2019}. Although most recent works rely on generalizable NeRFs like PixelNerf~\cite{pixelnerf}, MINE~\cite{mine2021}, or GRF~\cite{grf2021} which learn a representation generalizable to unseen input images.
The almost entire single-view literature however focuses on objects which hold specific challenges such as shape and appearance disentanglement~\cite{codenerf,sharf}, exploiting symmetry priors~\cite{symmnerf}, or category-centric/agnostic view synthesis~\cite{reizenstein2021common,visionnerf}.
In the latter, objects are usually on a plain background though CO3D~\cite{reizenstein2021common} handle objects on cluttered scenes or large-scale scenes being synthetic as in SEE3D~\cite{sharma2022seeing}, or real as in MINE~\cite{mine2021} or AutoRF~\cite{Mller2022AutoRFL3}. 
Specific to complex scenes, \cite{mine2021} synthesizes novel depths and views building on Multiplane Images, while very recently \cite{wimbauer2023behind} explored prediction of density fields trained with stereo or monocular sequences though getting limited improvement on the latter.

In general, \textbf{depth supervision} is shown to improve quality and convergence speed~\cite{dsnerf, Cao2022FWDRN, urbannerf, roessle2022depthpriorsnerf}, leveraging, for example, structure from motion~\cite{dsnerf,roessle2022depthpriorsnerf} or Lidar data~\cite{urbannerf}. Any NeRF-based method can implicitly optimize depth but those doing it explicitly still require depth supervision. Instead, we explicitly optimize depth \textit{self-supervisedly}.\\
Since NeRF optimizes radiance field only at sparse locations, \textbf{efficient sampling strategy} is needed to avoid prohibitive cost~\cite{donerf}. 
Departing from the initial hierarchical sampling~\cite{nerf}, a log warping strategy was proposed in DONeRF~\cite{donerf} with depth supervision, while~\cite{Kurz2022AdaNeRFAS} uses a pretrained NeRF, and~\cite{Kurz2022AdaNeRFAS} employs dual sampling-shading networks in a 4-stage training scheme.
We inspire from above works but approximates the continuous density volume as a mixture of Gaussians from which we can efficiently sample, without any complex setup.\\

\begin{figure*}[!t]
	\includegraphics[width=1.0\linewidth]{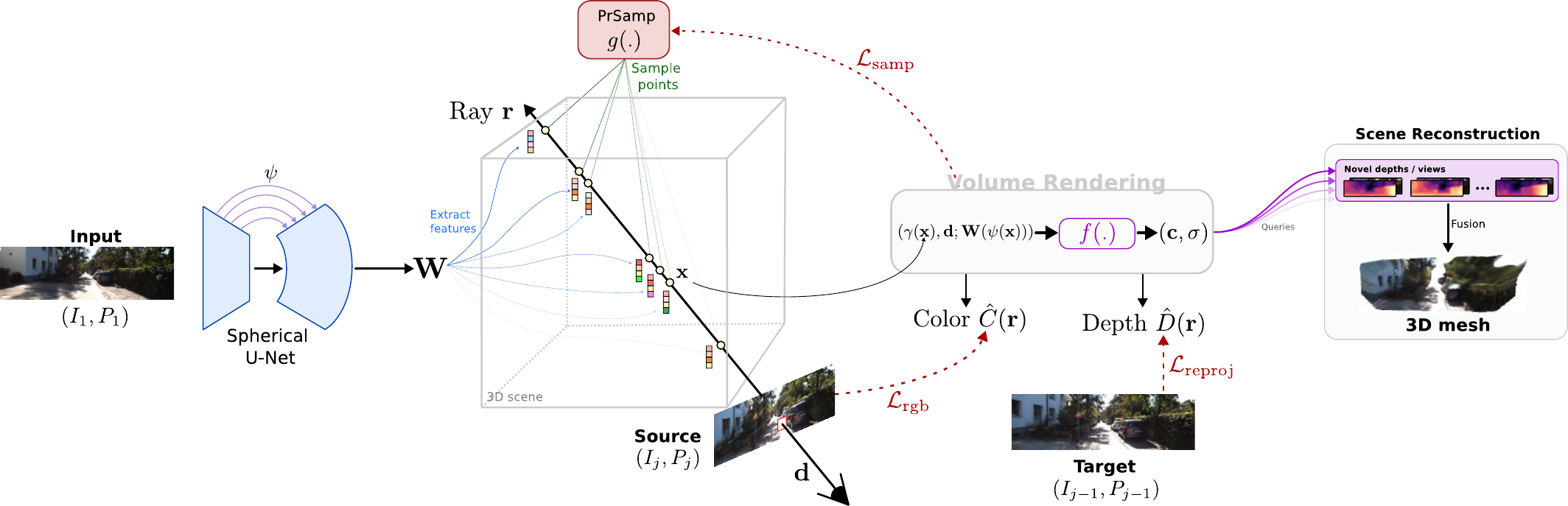}
	\caption{\textbf{Scene Representation Learning in SceneRF.} We leverage generalizable neural radiance field (NeRF) to generate novel depth views, conditioned on a single input frame. During training, for each ray $\mathbf{r}$ in addition to color $\hat{C}$, we explicitly optimize depth $\hat{D}$ with a reprojection loss $\mathcal{L}_\text{reproj}$ (\cref{sec:nerf_for_depth}), introduce a Probabilistic Ray Sampling strategy (PrSamp,~\cref{sec:ray_sampling}) to sample points more efficiently. To hallucinate features outside the input FOV, we propose a spherical U-Net (\cref{sec:sphere}). Finally, our scene reconstruction scheme (\cref{sec:reconstruction}) fuses novel views/depths to estimate the 3D mesh.}
	\label{fig:overview}
\end{figure*}
\paragraph*{3D reconstruction}

While early deep methods focused on reconstruction with explicit representations: ~like voxels~\cite{pix2voxplus}, point clouds~\cite{achlioptas2018poincloud, Fan_2017_CVPR,yang2019pointflow} or meshes~\cite{wang2018pixel2mesh,chen2020bspnet,deepmarchingcube}, recently, implicit representations gain popularity~\cite{Park2019DeepSDFLC, PengNMP020convoccnet,Popov2020CoReNetC3, Huang2022PlanesVC}. %
A common practice for 3D object reconstruction is to employ object detectors~\cite{izadinia2017im2cad, gkioxari2020meshRCNN,zhang2020perceiving,grabner2019location}. %
A number of works addressed holistic 3D scene understanding, seeking prediction of geometry and semantics for indoor~\cite{nie2020total3dunderstanding,huang2018holistic,zhang2021holistic,lee2017roomnet,sun2019horizonnet,zou2018layoutnet, dahnert2021panoptic, Ebrahimi2022DifferentiableGS} and outdoor scenes~\cite{Zakharov2021SingleShotSR}, or both~\cite{monoscene}. When semantic and geometry are estimated jointly it is referred as semantic scene completion (SSC), recently surveyed in~\cite{roldao20223d}. %
Relevant to this work, MonoScene~\cite{monoscene} and its descendants~\cite{miao2023occdepth,li2023voxformer, tpvformer} address SSC with single-input view but requiring 3D supervision.

A few alternatives exist for \textbf{self-supervised 3D reconstruction}. The straightforward use of monocular depth estimation, reviewed in~\cite{ming2021deep}, inherently limits reconstruction to the visible surface. 
Differentiable renderers are also popular, trained with views and poses~\cite{Niemeyer2020DifferentiableVR,sitzmann2019srns,Duggal2022TopologicallyAwareDF}. To alleviate the need of color rendering, some optimize silhouettes~\cite{Han2020DRWRAD} or 2D projection~\cite{Zubic2021AnEL}. 
Despite dazzling visuals, they remain object-centric.
Instead, we learn scene reconstruction self-supervisedly from a general radiance field.

\section{SceneRF}
\label{sec:method}
SceneRF learns the implicit scene geometry from a single monocular RGB image, training in a self-supervised manner with image-conditioned Neural Radiance Fields (NeRFs)~\cite{nerf,pixelnerf}. 
Given a set~$\mathcal{S}$ of image sequences with $m$ temporally consecutive RGB images with corresponding poses, denoted $\{(I^{s}_1, P^s_1), \dots, (I^s_{m}, P^s_{m})\}_{s\in\mathcal{S}}$, we learn a neural representation conditioned on the first frame of the sequence $\{I^{s}_1\}_{s\in\mathcal{S}}$.
The conditioning learned \textit{is shared across sequences} and self-supervisedly optimized with all other frames (\ie, $\{I^{s}_2, ..., I^{s}_m\}_{s\in\mathcal{S}}$). Subsequently, it can be used for 3D reconstruction from a single RGB image.

In~\cref{sec:nerf_for_depth} we elaborate on our usage of NeRF for novel depth synthesis relying on optimization with a reprojection loss. We then detail two major components.
First, in~\cref{sec:ray_sampling} we introduce a topology-preserving strategy to efficiently sample points close to the surface. 
Second, to hallucinate the scene \textit{beyond} the input image field of view, we introduce our custom U-Net~\cref{sec:sphere} with a spherical decoder.
Ultimately, the above design choices allow us to synthesize novel depth/views at arbitrary positions which are then fused into a single 3D reconstruction~\cref{sec:reconstruction}.

\subsection{NeRF for novel depth synthesis}
\label{sec:nerf_for_depth}
In their original formulation, NeRFs~\cite{nerf,pixelnerf} optimize a continuous volumetric radiance field $f(.) = (\sigma, \mathbf{c})$ such that for a given 3D point $\mathbf{x} \in \mathbb{R}^3$ and viewing direction $\mathbf{d} \in \mathbb{R}^3$, it returns a density $\sigma$ and RGB color $\mathbf{c}$. In the following, we build on PixelNeRF~\cite{pixelnerf} to learn a generalizable radiance field across sequences, and introduce new design choices to efficiently synthesize novel depth views.

The training of SceneRF is illustrated in \cref{fig:overview}. Given the first 
\textit{input}
frame ($I_1$) of a sequence\footnote{For clarity, we hereafter omit the superscript sequence $^s$, but the process applies to all training sequences $\mathcal{S}$.}, we extract a feature volume $\mathbf{W}=\textit{E}(I_1)$ with our SU-Net (\cref{sec:sphere}). 
We then select randomly a \textit{source} future frame $I_j, 2 \leq j \leq m$, and randomly sample $\ell$ pixels from it. 
Given known \textit{source} pose and camera intrinsics, we efficiently sample $N$ points along the rays passing through these pixels~(\cref{sec:ray_sampling}). 
Each sampled point $\mathbf{x}$ is then projected on a sphere with $\psi(\cdot)$ so we can retrieve the corresponding \textit{input} image feature vector $\mathbf{W}(\psi(\mathbf{x}))$ from bilinear interpolation. 
The latter is passed to the NeRF MLP~$f(\cdot)$, along with viewing direction $\mathbf{d}$ and positional encoding $\gamma(\mathbf{x})$, to predict the point density $\sigma$ and RGB color $\mathbf{c}$ in the input frame coordinates.
This writes:
\begin{equation}
	f(\gamma(\mathbf{x}), \mathbf{d}; \mathbf{W}(\psi(\mathbf{x}))) = (\mathbf{c}, \sigma)
	\label{eq:f}
\end{equation}
As in original NeRF~\cite{nerf}, quadrature approximates the color $\Cest$ of camera ray $\ray$ from colors sampled along the ray. For the sake of generality, we write it as:

\begin{equation}
	\Cest = \sum_{i}^{N} w_i \mathbf{c}_i\,\text{ where}\,	
	w_i = T_i (1 - \exp(-\sigma_i \delta_i)),
	\label{eq:Cest}
\end{equation}
with $T_i$ the accumulated transmittance and $\delta_i$ is the distance to the previous adjacent point, as defined in~\cite{nerf}. 

\subsubsection{Depth optimization}
\label{sec:depth_optim}
Unlike most NeRFs, we seek to unravel depth \textit{explicitly} from the radiance volume and therefore define its estimation $\Dest$ as:
\begin{equation}
	\Dest = \sum_{i}^{N} w_i d_i\,,
	\label{eq:Dest}
\end{equation}
where $d_i$ is the distance of point $i$ to the sampled position. 

To optimize depth {without ground-truth supervision}, we inspire from self-supervised depth methods~\cite{monodepth17, monodepth2}, and apply a photometric reprojection loss between the warped \textit{source} image $I_j$ and its preceding frame $I_{j-1}$, referred as \textit{target}. 
We choose consecutive frames to ensure maximum overlaps.
Using the sparse depth estimate $\hat{D}_j$, the photometric reprojection loss $\Lreproj$ writes:
\begin{equation}
	\Lreproj = \frac{1}{\ell}\sum_{i=1}^{\ell} ||I_{j}(i) - I_{j-1}\bigg(\proj\big(\hat{D}_j(i)\big)\bigg) ||_1\,,
	\label{eq:reproj}
\end{equation}
with $\proj(\cdot)$ the projection of 2D coordinates $i$ in $I_{j-1}$ using ad-hoc camera intrinsics and poses. Importantly, note that while $\hat{D}_j$ is sparse --- since only estimated for \textit{some} rays --- the stochastic nature of these rays offers statistically dense supervision. To also account for moving objects, we apply the pixels auto-masking strategy from~\cite{monodepth2}.

\subsection{Probabilistic ray sampling (PrSamp)}
\label{sec:ray_sampling}
\begin{figure}
	\centering
	\includegraphics[width=0.95\linewidth]{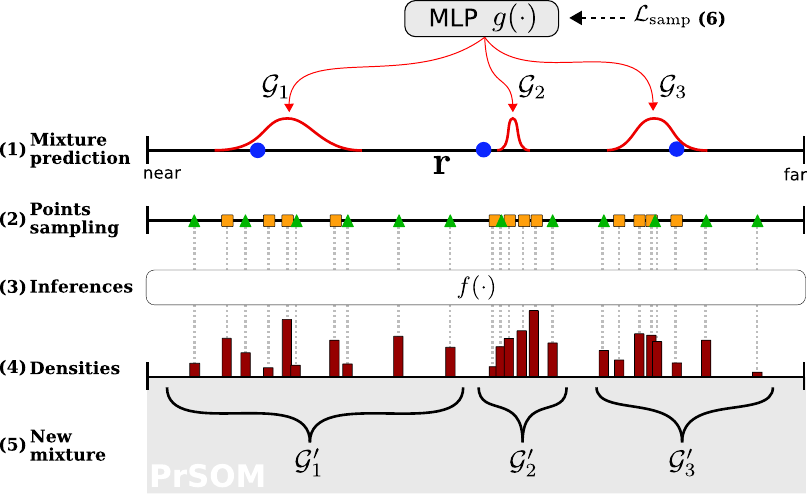}
	\caption{\textbf{Probabilistic Ray Sampling (PrSamp).} Here, $k{=}3$ Gaussians and $m{=}4$ points per Gaussian. Refer to \cref{sec:ray_sampling} for details.} %
	\label{fig:prsom}
\end{figure}
Prior works~\cite{donerf, hu2022efficientnerf, nerf} demonstrate that for volume rendering, sampling points \textit{close to the scene surface} improves color estimation (\ie, \cref{eq:Cest}) while reducing its computational cost due to less $f(\cdot)$ inferences.
This is however not trivial here since we lack depth guidance making surface location unknown.

To address this, our probabilistic ray sampling strategy (PrSamp) models the continuous density along each ray as a mixture of 1D Gaussians which then serve as support for points sampling. 
PrSamp implicitly learns to correlate high mixture values with surface locations, subsequently allowing better sampling with much less points. For example, optimization of a 100m volume requires only 64 points per ray.

\newcommand{\step}[1]{\textbf{(#1)}}

Referring to symbols and (\textbf{steps}) in \cref{fig:prsom}, for each ray~$\mathbf{r}$ we first uniformly sample $k$ points ($\textcolor{blue}\bullet$) between \textit{near} and \textit{far} bounds. 
\step{1}~Taking as input the points $\textcolor{blue}\bullet$ and their corresponding features $\mathbf{W}(\psi({\textcolor{blue}\bullet}))$, a dedicated MLP $g(\cdot)$ predicts a mixture of $k$ 1D Gaussians $\{\gauss_1,\dots,\gauss_k\}$.
\step{2}~We then sample $m$ points {per Gaussian}~($\textcolor{orange}\blacksquare$) and 32 more points uniformly~($\textcolor{green}\blacktriangle$)
; which amounts to $N{=}k{\times{}}m{\textcolor{orange}\blacksquare}{+}32{\textcolor{green}\blacktriangle}$ points. \textit{The addition of uniform points is essential to explore the scene volume and prevent $g(\cdot)$ from falling into local minima.}
\step{3}~All points are then passed to $f(\cdot)$ in \cref{eq:f} for volume rendering of color $\Cest$ and depth $\Dest$. 
\step{4}~Intuitively, the densities $\{\sigma_1,\dots,\sigma_N\}$ inferred by $f(\cdot)$ are cues for 3D surface locations, which we use to update our mixture of Gaussians. 
To solve the underlying points-Gaussians assignment problem \step{5}~we rely on Probabilistic Self-Organizing Maps~(PrSOM) from~\cite{prsom}. In a nutshell, PrSOM assigns points to Gaussians from the likelihood of the former to be observed by a set of points while strictly preserving the mixture topology. 
For each Gaussian $\gauss_i$ and its assigned points $\mathcal{X}_i$
, the {updated} $\gauss'_i$ is the average of all points $j \in \mathcal{X}_i$, weighted by the
conditional probability $p(j/\gauss_i)$ defined in~\cite{prsom} and the occupancy probability\footnote{We use alpha values from~\cite{nerf} as good-enough occupancy estimators: $\alpha_j{=}1{-}\exp({-}\sigma_j \delta_j)$ with $\delta_j$ the distance to previous point.} of~$j$.

\noindent{}Finally, \step{6} the Gaussians predictor $g(\cdot)$ is updated from the mean of KL divergences between the current and the new Gaussians:
\begin{equation}
	\mathcal{L}_\text{gauss} = \frac{1}{\ngauss}\sum_i^k \text{KL}(\gauss_i||\gauss'_i)\,.
\end{equation}
To further enforce one Gaussian \textit{on the visible surface}, we also minimize distance between depth and closest Gaussian:
\begin{equation}
	\mathcal{L}_\text{surface} = \min_i (||\mu(\gauss'_i) - \Dest||_1)\,.
\end{equation}
The complete loss is the sum: $\Lsamp=\mathcal{L}_\text{gauss}+\mathcal{L}_\text{surface}$. 

In practice, we use $k=4$ Gaussians and $m=8$ points per Gaussians, leading to only $N=64$ points per ray. The pseudo code is shown in~\cref{algo:prsamp}. We ablate parameters in~\cref{sec:exp_ablation}.

\subsection{Spherical U-Net (SU-Net)}
\label{sec:sphere}

By definition, the validity domain of $f(.)$ is restricted to the feature volume $\mathbf{W}(.)$ which for a standard U-Net is the camera FOV, thus preventing estimation of color and depth (Eqs.~\ref{eq:Cest},\ref{eq:Dest}) outside of the FOV where features cannot be extracted. This is unsuitable for scene reconstruction.

Instead, we equip our SU-Net with a decoder convolving in the spherical domain. Because spherical projection induces less distortion than its planar counterpart~\cite{salomon2006transformations} we may enlarge the FOV (typically, approx.~$120^\circ$) to hallucinate color and depth beyond the input image FOV.

At the bottleneck, the encoder features are mapped to an arbitrary sphere with $\psi(.)$ and passed to our spherical decoder. 
To cope with wide feature space at low cost, we employ light-weight dilated convolutions in the spherical decoder
and adapt the standard U-Net multi-scale skip-connections simply by mapping features with $\psi(.)$.

In practice, we map a 2D pixel $[\xpix, \ypix]^\top$ to its \textit{normalized} latitude-longitude spherical coordinates $[\theta, \phi]$. Considering $\begin{bmatrix} \xray, \yray, 1\end{bmatrix}^\top{\sim}\text{K}^{-1}
\begin{bmatrix} \xpix, \ypix, 1\end{bmatrix}^\top$ a ray passing through said pixel and the camera center. The projection writes:
\begin{equation}
	\psi\begin{pmatrix}
		x  \\
		y
	\end{pmatrix} {=}
	\begin{pmatrix}
		\theta  \\
		\phi
	\end{pmatrix}
	{=}
	\begin{pmatrix}
		\pi-\arctan(\xray^{-1}) \\
		\arccos(-\yray/r)
	\end{pmatrix}
	\label{eq:sphere_mapping}
\end{equation}
where $r = \sqrt{\xray^2 + \yray^2 + 1}$. 
When inputted in the decoder, $[\theta, \phi]$ are discretized uniformly and features stored in a tensor that covers an arbitrary large FOV.

\subsection{Scene reconstruction scheme}
\label{sec:reconstruction}

With prior sections, SceneRF is now equipped with novel depth synthesis capability that allows us to synthesize depth that significantly diverges from the source input position. 
We use this ability to frame scene reconstruction as the composition of multiple novel depth views. 

\begin{figure}
	\centering
	\includegraphics[width=0.9\linewidth]{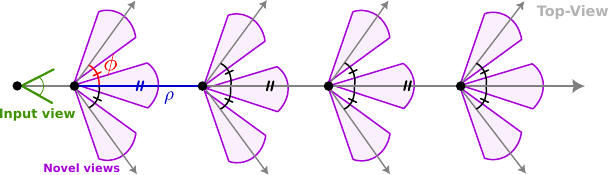}
	\caption{\textbf{Reconstruction scheme.} Given an input image, we fuse the TSDF of the synthesized novel depth/views uniformly sampled along an imaginary path, at varying angles.}%
	\label{fig:reconstruction}
\end{figure}

As illustrated in Fig.~\ref{fig:reconstruction}, given an input frame we synthesize novel depths along an imaginary straight path, uniformly every $\rho$ meters up to a given distance. At each position, we also vary the horizontal viewing angles $\Phi{=}\{-\phi,0,\phi\}$. 

The synthesized depths are then converted to TSDF using~\cite{3dmatch} and the overall scene TSDF for voxel $v$ is obtained from the minimum of all: $\text{V}(v) = \text{TSDF}_{\text{argmin}_i |\text{TSDF}_i(v)|}(v)$, where $i$ spans all synthesized depths.
Traditionally, a voxel TSDF is the weighted average of all TSDFs~\cite{oldtsdffusion, kinectfusion}, but we empirically show (see \cref{supp:sec:recon_detail}) that using the minimum leads to better results. We conjecture that this relates to the linearly increasing depth error with distance.

\section{Experiments}
\label{sec:experiments}
We evaluate SceneRF on two primary tasks, namely novel depth synthesis and scene reconstruction, and novel view synthesis which we refer as `subsidiary task' because it is not used for scene reconstruction. 
While we do \textit{not} use 3D data, we need it for evaluation, and thus report results on SemanticKITTI~\cite{semkitti, kitticvpr} and BundleFusion~\cite{dai2017bundlefusion} for all three tasks.
Each dataset holds unique challenges. SemanticKITTI has large driving scenes ($\approx$100m deep) and the image sequences are captured from a forward-facing camera which offers little viewpoint variations. Instead, BundleFusion has shallow indoor scenes ($\approx$10m) with sequences exhibiting large lateral motion.
Since we first address \textit{self-supervised} monocular scene reconstruction from RGB images, we detail our non-trivial adaptation of monocular reconstruction baselines~\cite{monoscene, 3DSketch, aicnet} (\cref{sec:baselines}). 

We always use $k=4$ gaussians and $m=8$ points per Gaussians in PrSamp (\cref{sec:ray_sampling}) but vary novel depth/view sampling for reconstruction (\cref{sec:reconstruction}). Specifically, we sample views every $\rho=0.5$m for up 10m at angles $\Phi=\{-10, 0, +10\}$ for SemanticKITTI, and use $\rho=0.2$m for up to 2.0m with $\Phi=\{-20, 0, +20\}$ for BundleFusion.

\begin{table*}[!t]
	\scriptsize
	\centering
	\setlength{\tabcolsep}{0.005\linewidth}
	\resizebox{\linewidth}{!}{
		\begin{tabular}{c|ccccccc|ccc|ccccccc|ccc}
			\toprule
			\multirow{2}[2]{*}{Method} & \multicolumn{10}{c|}{\small\textbf{SemanticKITTI}} & \multicolumn{10}{c}{\small\textbf{BundleFusion}} \\
			\multirow{2}[2]{*}{} & \multicolumn{7}{c|}{\textbf{Novel depth synthesis}} & \multicolumn{3}{c|}{{Novel view synthesis}} & \multicolumn{7}{c|}{\textbf{Novel depth synthesis}} & \multicolumn{3}{c}{{Novel view synthesis}} \\
			& Abs Rel$\downarrow$ & Sq Rel$\downarrow$ & RMSE$\downarrow$ & RMSE log$\downarrow$ & $\delta$1$\uparrow$ & $\delta$2$\uparrow$  & $\delta$3$\uparrow$  & LPIPS$\downarrow$ & SSIM$\uparrow$  & PSNR$\uparrow$  & Abs Rel$\downarrow$ & Sq Rel$\downarrow$ & RMSE$\downarrow$ & RMSE log$\downarrow$ & $\delta$1$\uparrow$  & $\delta$2$\uparrow$  & $\delta$3$\uparrow$  & LPIPS$\downarrow$ & SSIM$\uparrow$  & PSNR$\uparrow$  \\
			\midrule
			MonoDepth2~\cite{monodepth2} & 0.5259 & 7.113 & 14.43 & 1.0292 & 10.44 & 26.32 & 41.43 & 0.623 & 0.166 & 9.61  & 0.3205 & 0.562 & 0.879 & 0.4080 & 44.98 & 76.31 & 91.05 & 0.537 & 0.492 & 11.15  \\
			MonoDepth2 + LaMa~\cite{lama} & 0.4086 & 5.101 & 12.14 & 0.8472 & 30.93 & 49.50 & 62.65 & 0.489 & 0.418 & 15.32 & 0.3937 & 0.954 & 1.155 & 0.4538 & 46.43 & 75.10 & 88.79 &  0.338 & 0.794 & 20.80 \\
			SynSin~\cite{synsin} &   0.3611 & 3.483 & 8.824 & 0.4290 & 52.61 & 74.56 & 86.50 & 0.519 & 0.375 & 14.86 & 0.2360 & 0.174 & 0.522 & 0.2992 & 57.08 & 84.71 & 95.53 & 0.627 & 0.597 & 13.48  \\
			PixelNeRF~\cite{pixelnerf}&   0.2364 & 2.080 & 6.449 & 0.3354 & 65.81 & 85.43 & 92.90 & 0.489 & 0.466 & 15.80  & 0.6029 & 2.312 & 1.750 & 0.5904 & 46.34 & 72.38 & 83.89 &  0.351  & 0.822 &    20.51 \\
			MINE~\cite{mine2021} & 0.2248 & 1.787 & 6.343 & 0.3283 & 65.87	& 85.52 & 93.30 &  \textbf{0.448}  &  \textbf{0.496} &  16.03  &\underline{0.1839}	& \underline{0.098} & \underline{0.386} & \underline{0.2386} & \underline{65.53} & \underline{91.78} & \underline{98.21}  &  0.377 & 0.763 & \underline{20.60} \\
			VisionNerf~\cite{visionnerf} & \underline{0.2054} & \underline{1.490} & \underline{5.841} & \underline{0.3073} & \underline{69.11} & \underline{88.28} & \underline{94.37} & \underline{0.468} & \underline{0.483} & \textbf{16.49} & 0.5958 & 2.468 & 1.783 & 0.5586 & 55.47 & 79.29 & 86.68 & \underline{0.332} & \underline{0.831} & 20.51 \\
			SceneRF  & \textbf{0.1681}	& \textbf{1.291} & \textbf{5.781} & \textbf{0.2851} & \textbf{75.07} & \textbf{89.09} & \textbf{94.50} & 0.476 & 0.482 & \underline{16.46} & \textbf{0.1766}& \textbf{0.094} & \textbf{0.368} & \textbf{0.2100} & \textbf{72.71} & \textbf{94.89} & \textbf{99.23} &  \textbf{0.323} & \textbf{0.853} & \textbf{25.07} \\
			\bottomrule
		\end{tabular}
	}
	\caption{\textbf{Novel depth/view synthesis.} We outperform all on our main task of \textit{novel depth}, and perform on par on the subsidiary \textit{novel view} task. Note the large $\delta{}1$ gaps, in particular w.r.t. PixelNerf from which we depart from. (val. sets)}
	\label{tab:novelviews}
\end{table*}

\paragraph*{Datasets.} 
\textbf{SemanticKITTI}~\cite{semkitti} has pairs of outdoor geolocalized images with voxelized lidar scans of 256x256x32 with 0.2m voxel, with free/occupy labels. 
We use the standard train/val split as in~\cite{monoscene, semkitti} and left-crop RGB images to 1220x370. 
{We train SceneRF with successive frames spanning $\approx$10m while ensuring a minimum of 0.4m distance between two frames.}
This results in 10,270 training sequences. 
{We evaluate novel view at 1:3 resolution and novel depth at 1:2 against sparse lidar projection.}\\
\textbf{BundleFusion}~\cite{dai2017bundlefusion} has indoor scenes captured with a handheld device. It has RGB-D images of $640{\times{}}480$ each with an estimated 6-DOF pose. 
We drop every other frame to increase diversity, \ie getting 9733 images split in sequences of 17 frames. The middle frame serves as input and remaining ones for supervision. We select 7 of the 8 scenes for training and 1 as validation. We evaluate at 1:2 resolution.

\paragraph*{Metrics.} To measure our reconstruction quality, we use the intersection over union (IoU), precision, and recall of occupied voxels. 
For novel depth estimation, we choose usual metrics~\cite{monodepth2}: relative error absolute (Abs~Rel) or squared (Sq~Rel), root mean squared error (RMSE), mean $\log_{10}$ error (RMSE log), threshold accuracies ($\delta$1, $\delta$2, $\delta$3). As a common practice, depth is capped to 80m in SemanticKITTI and 10m in BundleFusion. 
Following~\cite{mine2021}, we measure the quality of synthesized RGB images with: Structural Similarity Index (SSIM)~\cite{ssim}, PSNR, and LPIPS perceptual similarity~\cite{lpips}.

\paragraph*{Training setup.} SceneRF trains end-to-end minimizing \makebox{$\mathcal{L}_{\text{total}} = \Lrgb + \Lreproj + \Lsamp$} 
where $\Lrgb$ is the standard L2 photometric reconstruction loss of NeRFs~\cite{urbannerf, nerf, pixelnerf}.
We report results for 50 epochs training with batch size of 4 and initial learning rate of 1e-5 with exponential decay at each epoch with gamma $0.95$.
Training was conducted on 4 Tesla v100 GPUs, amounting to ${\approx}5$~days.

\subsection{Baselines}
\label{sec:baselines}
\noindent\textbf{Novel depth/views.} Despite the bustling NeRF field, there are in fact few \textit{single-view} NeRFs. We select 3 of them among the best open-sourced ones for novel depths/views synthesis: PixelNeRF~\cite{pixelnerf}, VisionNeRF~\cite{visionnerf}, MINE~\cite{mine2021}. Similar to us, all train with images and poses. 
We also compare against state-of-the-art 3D-aware GAN, namely SynSin~\cite{synsin} {for which novel depths are obtained by applying its depth regressor on novel views.}
Finally, to account for natural baselines we evaluate against monocular depth estimation, here MonoDepth2~\cite{monodepth2}, where novel depths (views) are the reprojection of the (colored) point cloud derived from the input view and estimated depth map. As such novel views/depths are inevitably sparse we also report `MonoDepth2 + LaMa' where novel views of MonoDepth2 baseline are inpainted with LaMa~\cite{lama} and novel depth is obtained from running MonoDepth2 again\footnote{Empirically, we observe that directly depth inpainting is much worse.}.

\noindent\textbf{Scene reconstruction.} For monocular scene reconstruction, we consider 4 baselines being: MonoScene~\cite{monoscene}, LMSCNet$^{\text{rgb}}$~\cite{lmscnet},  3DSketch$^{\text{rgb}}$~\cite{3DSketch}, AICNet$^{\text{rgb}}$~\cite{aicnet}. 
The baselines with $^{\text{rgb}}$ are \textit{RGB-inferred version} from~\cite{monoscene}.
Since all baselines require geometric supervision from depth sensors, we report `3D' and `Depth' supervision along our `Image' supervision.
This is further detailed in~\cref{sec:exprecons}. 

\begin{table}
	\scriptsize
	\centering
	\setlength{\tabcolsep}{0.011\linewidth}
	{
		\begin{tabular}{c|c|c|c|ccc|ccc}
			\toprule
			\multirow{2}[1]{*}{Method} & \multicolumn{3}{c|}{\textbf{Supervision}} & \multicolumn{3}{c|}{\textbf{SemanticKITTI}} & \multicolumn{3}{c}{\textbf{BundleFusion}} \\
			&  3D & Depth & Image & IoU & Prec. & Rec. & IoU & Prec. & Rec. \\
			\midrule
			MonoScene~\cite{monoscene} & \checkmark & & & 37.14 & 49.90 & 59.24 & 30.15  & 35.07 & 68.51 \\ 
			\midrule 
			LMSCNet$^{\text{rgb}}$~\cite{lmscnet} & & \checkmark & & 12.08 & 13.00 & 63.16 & 14.91 & 25.22 & 31.15 \\ 
			3DSketch$^{\text{rgb}}$~\cite{3DSketch} & & \checkmark &  & 12.01 & 12.95 & 62.31 & 16.88 & 25.82 & 32.76 \\
			AICNet$^{\text{rgb}}$~\cite{aicnet}  & & \checkmark & & 11.28 & 11.84 & 70.89 & 15.99 & 25.20 & 30.41 \\ 
			MonoScene~\cite{monoscene} &  & \checkmark & & 13.53 & 16.98 & 40.06 & 19.00 & 22.51 & 54.91 \\ 
			\midrule 
			
			MonoScene*~\cite{monoscene}  & &  & \checkmark & 11.18 &  13.15 & 40.22 & 17.20 & 21.88 & 44.59 \\
			SceneRF & && \checkmark & \textbf{13.84} & \textbf{17.28} & \textbf{40.96}  & \textbf{20.16} & \textbf{25.82} & \textbf{47.92}\\
			\bottomrule
		\end{tabular}
		{\footnotesize * Here, MonoScene is supervised by depth predictions of~\cite{monodepth2} trained with ground-truth poses.}
	}
	\caption{\textbf{Scene reconstruction.} Despite being the \textit{only} self-supervised method, we outperform all `Depth' supervised baselines. Refer to~\cref{sec:exprecons} for supervision details.
	}
\label{tab:scenerecons}
\end{table}

\begin{table*}[!t]
	\scriptsize
	\centering
	\setlength{\tabcolsep}{0.005\linewidth}
	\resizebox{1.0\linewidth}{!}{
		\begin{tabular}{c|ccccccc|ccc|ccccccc|ccc}
			\toprule
			\multirow{2}[2]{*}{Method} & \multicolumn{10}{c|}{\small\textbf{SemanticKITTI}} & \multicolumn{10}{c}{\small\textbf{BundleFusion}} \\
			\multirow{2}[2]{*}{} & \multicolumn{7}{c|}{\textbf{Novel depth synthesis}} & \multicolumn{3}{c|}{{Novel view synthesis}} & \multicolumn{7}{c|}{\textbf{Novel depth synthesis}} & \multicolumn{3}{c}{{Novel view synthesis}} \\
			 & AbsRel$\downarrow$ & SqRel$\downarrow$ & RMSE$\downarrow$ & RMSElog$\downarrow$ & $\delta$1$\uparrow$ & $\delta$2$\uparrow$ & $\delta$3$\uparrow$ & LPIPS$\downarrow$ & SSIM$\uparrow$ & PSNR$\uparrow$ & AbsRel$\downarrow$ & SqRel$\downarrow$ & RMSE$\downarrow$ & RMSElog$\downarrow$ & $\delta$1$\uparrow$ & $\delta$2$\uparrow$ & $\delta$3$\uparrow$ & LPIPS$\downarrow$ & SSIM$\uparrow$ & PSNR$\uparrow$ \\
			\midrule 
			SceneRF & \best{0.1681}	& \best{1.291} & \best{5.781} & \best{0.2851} & \best{75.07} & \best{89.09} & \best{94.50} & \second{0.476} & \best{0.482} & \best{16.46} &\best{0.1766} & \second{0.094} & \best{0.368} & \second{0.2100} & \best{72.71} & \second{94.89} & \second{99.23} &  \second{0.323} & \best{0.853} & \second{25.07} \\
			w/o $\mathcal{L}_{\text{rgb}}$ & 0.1801 & 1.480 & 6.347 & 0.3085 & 72.15 & 87.56 & 93.66 & - & - & -& \second{0.1769} & \best{0.084} & \second{0.374} & \best{0.2043} & \second{71.75} & \best{95.82} & \best{99.79} & - & - & - \\
			w/o $\mathcal{L}_{\text{reproj}}$ & 0.2115 & 1.706 & 6.133 & 0.3059 & 69.10 & 87.55 & 94.13 & 0.491 & \second{0.481} & 16.42 & 0.2168 & 0.144 & 0.454 & 0.2577 & 64.99 & 90.47 & 97.72 & 0.328 & \second{0.852} & 24.82 \\
			w/o SU-Net & \second{0.1758} & 1.386 & 5.908 & 0.2967 & \second{73.91} & 88.27 & 94.01 &	\best{0.464} & 0.480 & 16.40 & 0.2449 & 0.167 & 0.488 & 0.3263 & 59.77 & 85.84 & 94.63 & 0.461 & 0.730 & 14.29 \\
			w/o PrSamp & 0.1858 &  \second{1.301} & \second{5.844} & \second{0.2936} & 71.85 & \second{88.73} & \second{94.24} & 0.505 & 0.471 & \second{16.43} & 0.1825 & 0.100 & 0.385 & 0.2125 & 70.69 & 94.10 & 98.78 &  \best{0.317} & 0.730 & \best{25.15} \\
			Freeze $\sigma$ $\mathcal{L}_\text{rgb}$ & 0.1750 & 1.366 & 6.029 & 0.2962 & 73.42 & 88.28 & 94.14 & 0.494 & 0.476 & 16.42 & 0.2081 & 0.131 & 0.423 & 0.2362 & 67.55 & 92.68 & 98.42 & 0.342 & 0.850 & 24.80 \\
			$\Lrgb$ on S + T & 0.1966 & 1.484 & 5.993 & 0.2991& 70.36 & 88.35 & 94.07 & 0.486 & 0.478 &	16.40 & 0.1942 & 0.134 & 0.409 & 0.2270 & 70.78 & 93.73 & 98.18 & 0.357 & 0.838 & 24.71 \\
			\bottomrule
		\end{tabular}
	}
	\caption{\textbf{Architecture ablation on the validation set.} All components contribute to yielding better results for our primary task of \textit{novel depth synthesis}, with mixed results on \textit{novel view synthesis}. Details are in~\cref{sec:exp_ablation}.}
	\label{tab:arch_ablation}
\end{table*}

\subsection{Novel depth synthesis}

To first evaluate the quality of our novel depths/views, given an input image we synthesize depth/views at the position of 
all frames in the sequence except for the input one. 

From \cref{tab:novelviews}, for the task of \textit{novel depth synthesis} we rank first on all metrics with a comfortable margin. %
In particular, one may note the large gaps on AbsRel and \makebox{$\delta{}$-metrics} as they are challenging metrics.
It is also noticeable that we significantly improve over PixelNeRF, from which we depart, demonstrating the benefit of our design choices.
For example, we get an improvement of +9.26 and +26.37 for $\delta{}1$ on SemanticKITTI and BundleFusion, respectively, w.r.t. PixelNeRF. Unsuprisingly, we outperform very significantly baselines using monocular depth estimation (\ie, MonoDepth2) or 3D-GAN (\ie, SynSin) which we ascribe to radiance volumes preserving 3D-aware consistency.

Though of least importance for scene reconstruction, \cref{tab:novelviews} also shows that SceneRF is roughly on par with the best methods on the subsidiary task of \textit{novel views synthesis} %
where, notably, we always improve over PixelNeRF.

In \cref{fig:qualitative_kitti}, we primarily show novel depths and the subsidiary novel views for varying input frames, multiple positions and angles w.r.t. the input frame position. 
For all, novel depths are visually outperforming the baselines. In particular, we note the sharper depth edges and the better quality at far when zooming in. When varying the viewing angle (\ie, $-10^\circ$ or $+10^\circ$) we note also fewer edge artefacts than baselines, which is even more striking for the outdoor example. Please also refer to the supplemental video.

\subsection{3D reconstruction results}
\label{sec:exprecons}
To evaluate reconstruction, we compare against the voxelized 3D groundtruth which is obtained either from the accumulation of lidar scans in SemanticKITTI or the fusion of depth maps in BundleFusion. 

Though we do not require depth or 3D for supervision, we still report 3 supervision setups in~\cref{tab:scenerecons}: (i)~`\textbf{3D}'~where baselines are trained with full 3D groundtruth. (ii)~`\textbf{Depth}' using as supervision the TSDF fusion~\cite{3dmatch} of depth sequences from the \textit{supervised} AdaBins method~\cite{adabin} which we retrain to boost performance. (iii)~`\textbf{Image}' where like in SceneRF, we only train self-supervisedly from image sequences. It is important to note that, except for the 'Image'-supervision baseline, all other baselines incorporate some sense of ground truth depth which we do not have. 

From~\cref{tab:scenerecons}, SceneRF is the only original self-supervised baseline that still \textit{outperforms all `{Depth}'-supervised baselines} on both datasets.
This is surprising given the additional geometrical supervision of `Depth' methods. It advocates that SceneRF efficiently self-discovers geometrical cues from image sequences. 
For more in depth comparison, we also adapt MonoScene~\cite{monoscene} to `Image'-supervision, using as ground truth the fusion of depth predictions of~\cite{monodepth2}\footnote{We train Monodepth2~\cite{monodepth2} with groundtruth poses for fair comparison.}. SceneRF still outperforms this image-supervised MonoScene by $\approx$ 3 points on BundleFusion. We also report the original `3D'-supervised MonoScene, acting as an unreachable upper bound since 3D provides supervision beyond occlusions.
In general, The low numbers for 'Depth' and 'Image' methods suggest task complexity, indicating potential for future research.

\begin{figure*}[!t]
\centering
\newcolumntype{P}[1]{>{\centering\arraybackslash}m{#1}}
\setlength{\tabcolsep}{0.004\textwidth}
\renewcommand{\arraystretch}{0.8}
\footnotesize
\begin{tabular}{P{0.12\textwidth}  P{0.10\textwidth}  P{0.18\textwidth}  P{0.14\textwidth} P{0.14\textwidth} P{0.14\textwidth}  P{0.14\textwidth}}		
	\multirow{2}{*}{Input}  & \multirow{2}{*}{Method}  & \multirow{2}{*}{\makecell{3D mesh\\(w/ our recons. scheme)}}  &  \multicolumn{3}{c}{Novel depth}     &  {Novel view} \\
	\cmidrule{4-6}%
	& & & $+1$m, $0^{\circ}$ & $+3$m, ${-}10^{\circ}$ & ${+}5$m, ${+}10^{\circ}$ & ${+}5$m, ${+}10^{\circ}$ \\[-0.1em]
	
	\multirow{4}{*}{\vspace{-10em}\includegraphics[width=\linewidth]{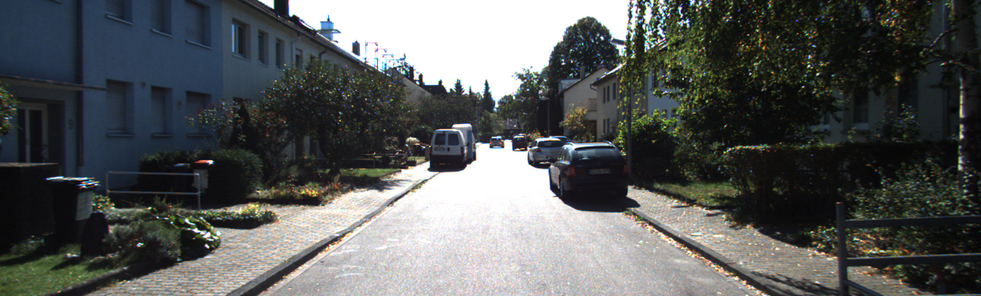}}& PixelNeRF & \includegraphics[width=\linewidth]{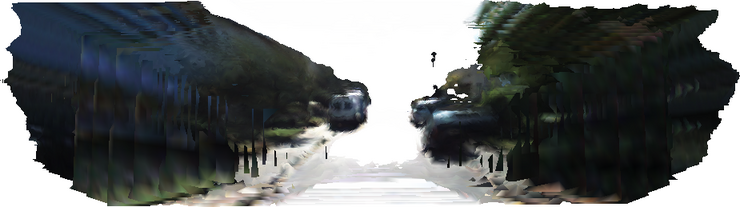} &
	\includegraphics[width=\linewidth]{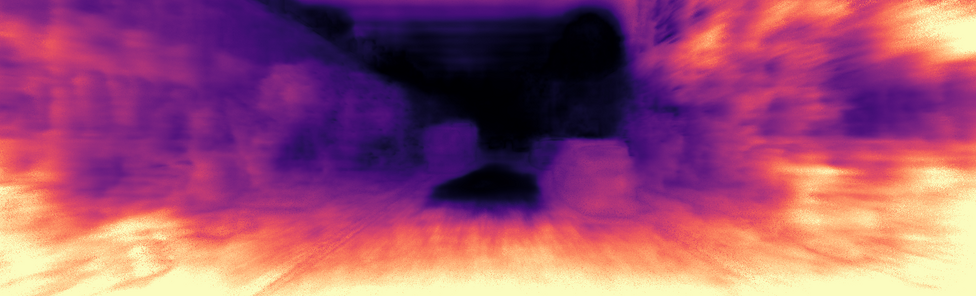}  &
	\includegraphics[width=\linewidth]{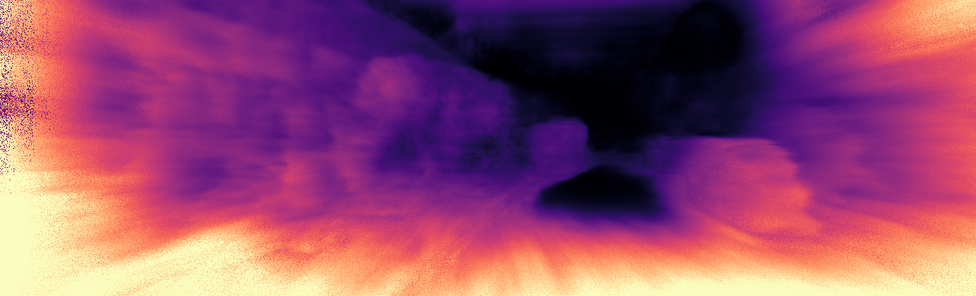}&
	\includegraphics[width=\linewidth]{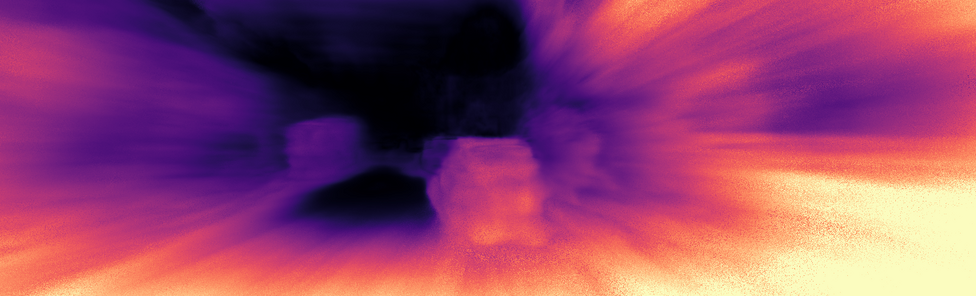} &
	\includegraphics[width=\linewidth]{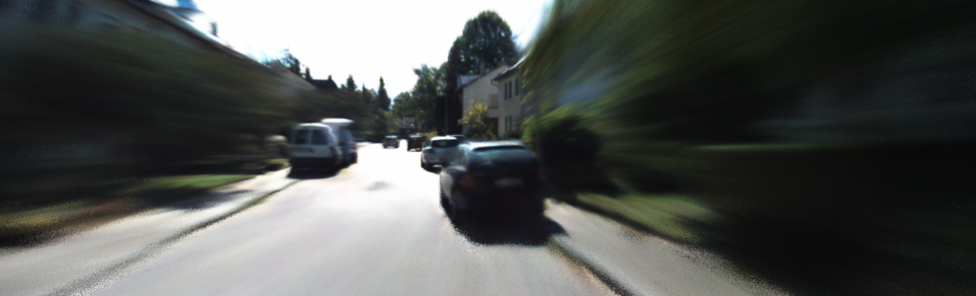} \\
	& MINE & \includegraphics[width=\linewidth]{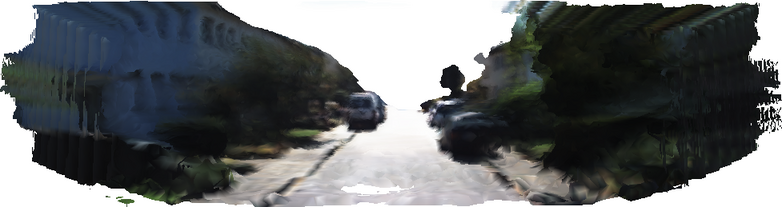}&
	\includegraphics[width=\linewidth]{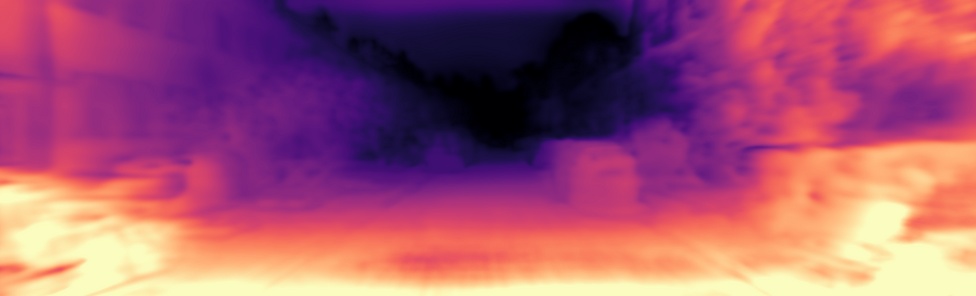}  &
	\includegraphics[width=\linewidth]{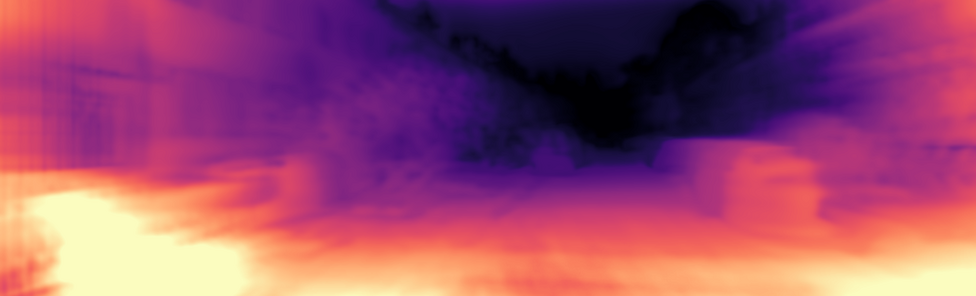}&
	\includegraphics[width=\linewidth]{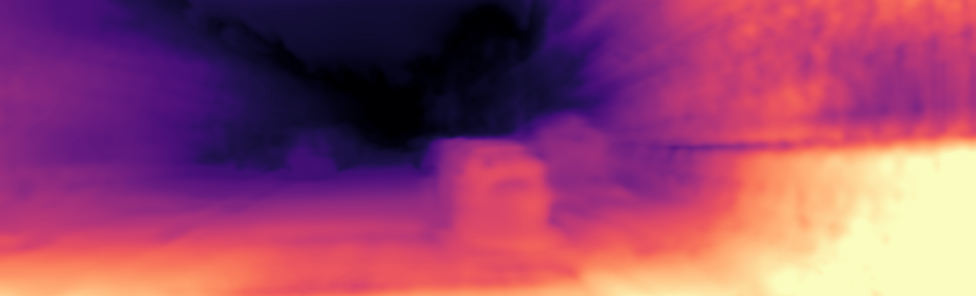} &
	\includegraphics[width=\linewidth]{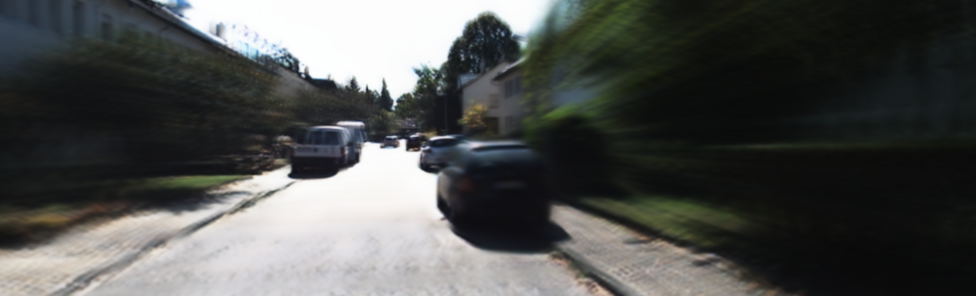}\\
	& VisionNeRF & \includegraphics[width=\linewidth]{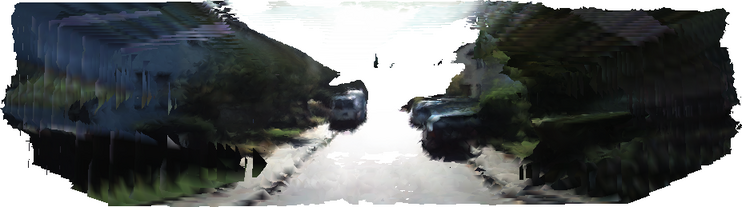}&
	\includegraphics[width=\linewidth]{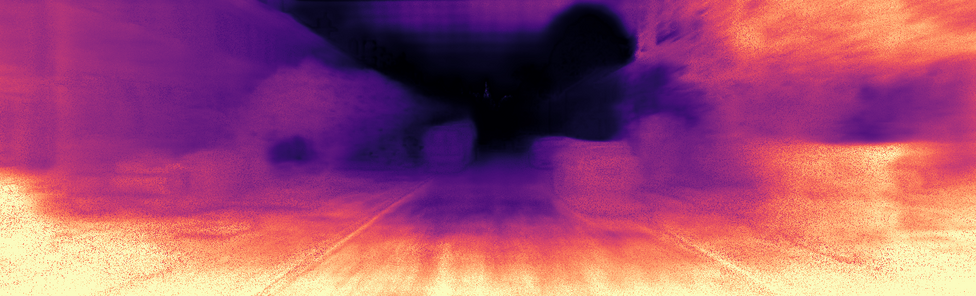}  &
	\includegraphics[width=\linewidth]{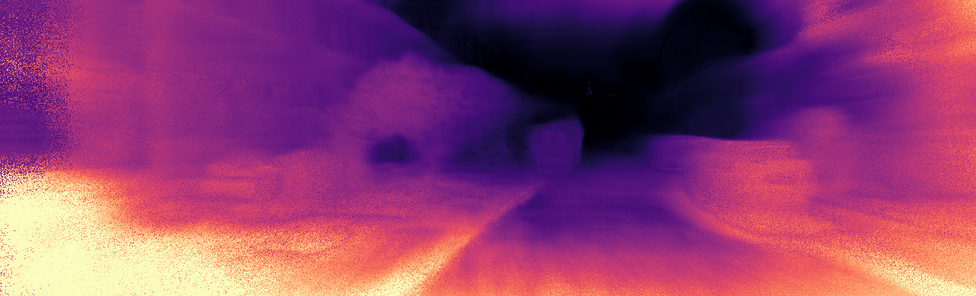}&
	\includegraphics[width=\linewidth]{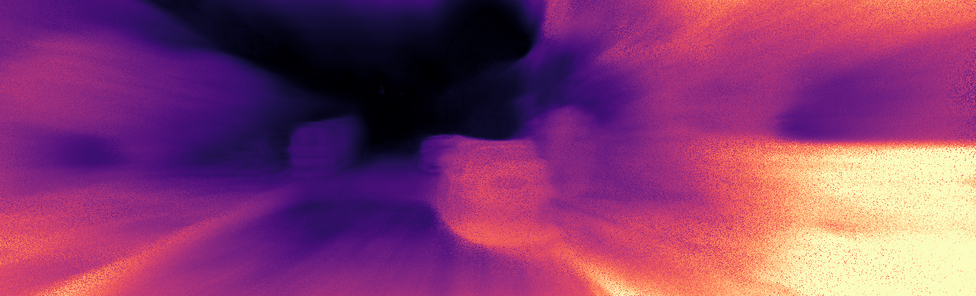} &
	\includegraphics[width=\linewidth]{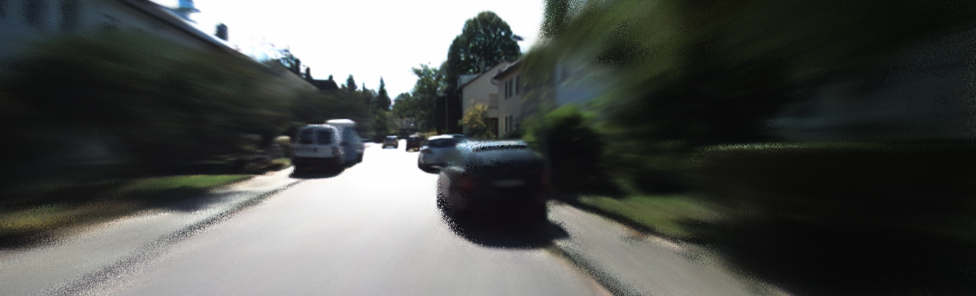}\\
	 & \textbf{SceneRF} & \includegraphics[width=\linewidth]{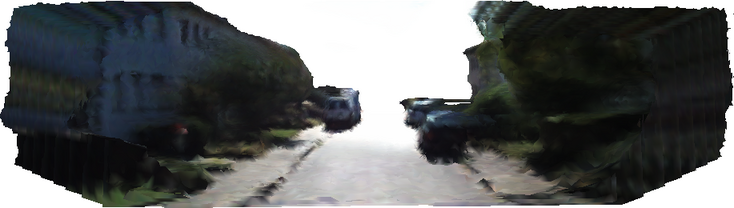}&
	\includegraphics[width=\linewidth]{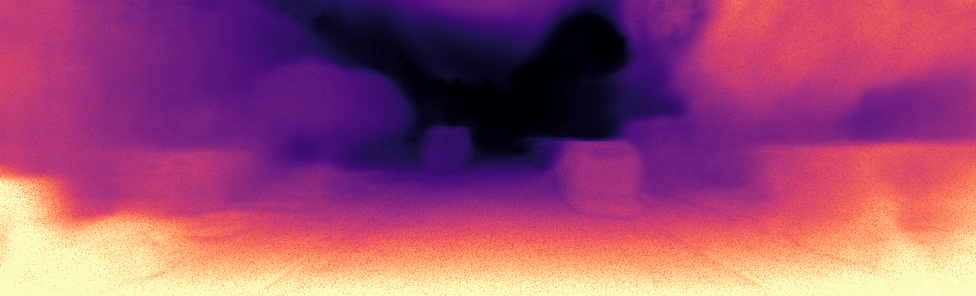}  &
	\includegraphics[width=\linewidth]{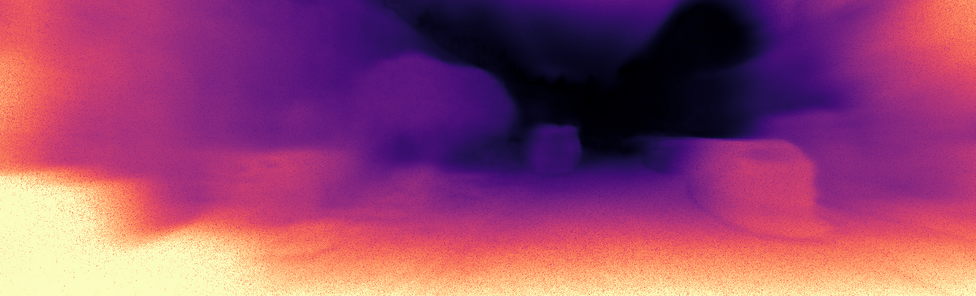}&
	\includegraphics[width=\linewidth]{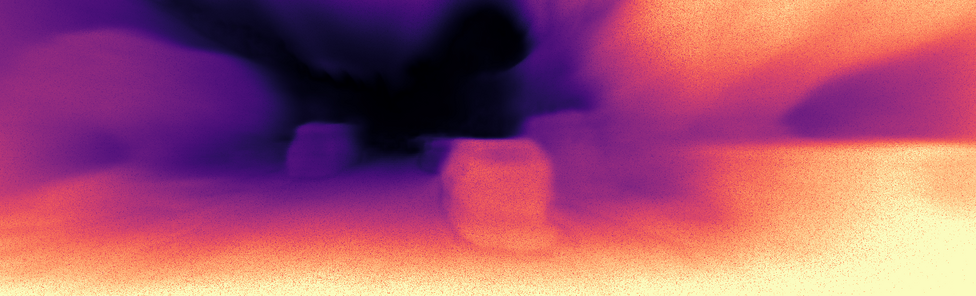} &
	\includegraphics[width=\linewidth]{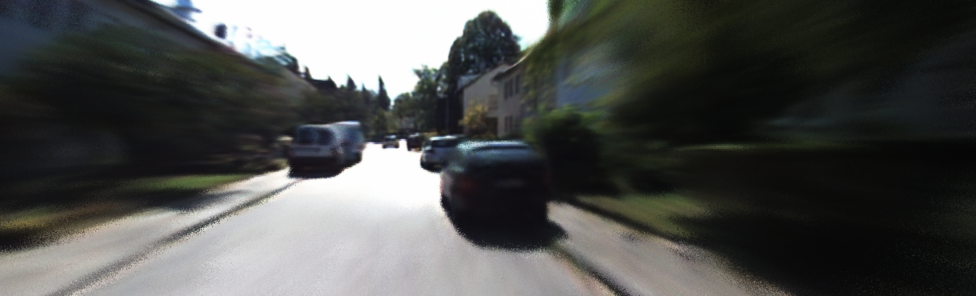} \\
	
	\arrayrulecolor{gray!50}
	\midrule

		& & & $+0.2$m, $0^{\circ}$ & $+0.2$m, ${-}20^{\circ}$ & ${+}0.4$m, ${+}20^{\circ}$ & ${+}0.4$m, ${+}20^{\circ}$ \\[-0.7em]
	
	\multirow{4}{*}{\vspace{-15em}\includegraphics[width=\linewidth]{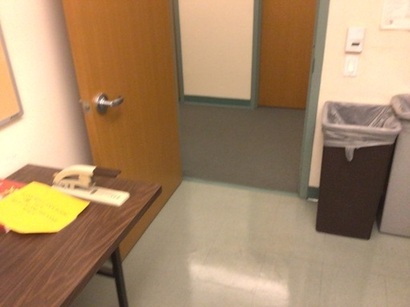}} & PixelNeRF & \includegraphics[width=\linewidth]{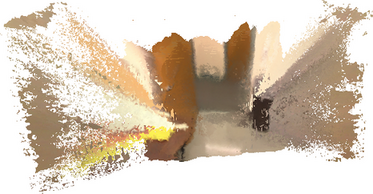} &
	\includegraphics[width=0.65\linewidth]{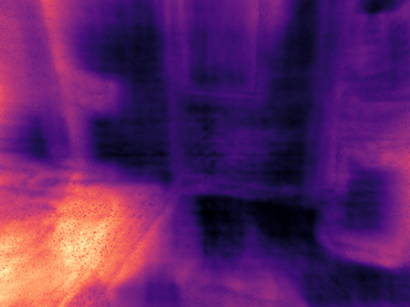} &
	\includegraphics[width=0.65\linewidth]{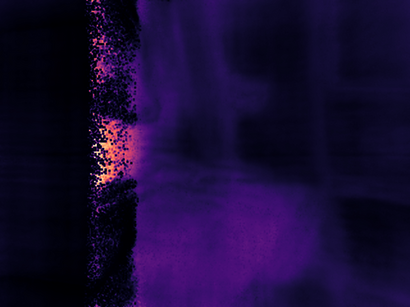} &
	\includegraphics[width=0.65\linewidth]{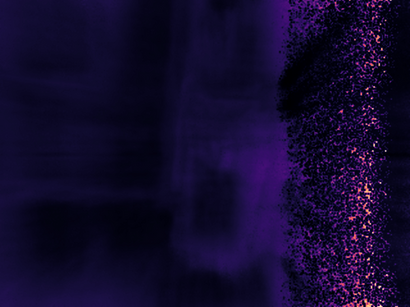} &
	\includegraphics[width=0.65\linewidth]{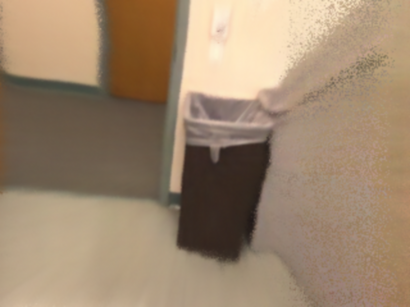} \\
	& MINE & \includegraphics[width=\linewidth]{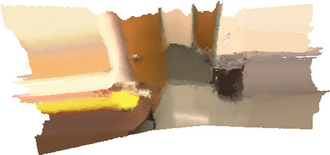} &
	\includegraphics[width=0.65\linewidth]{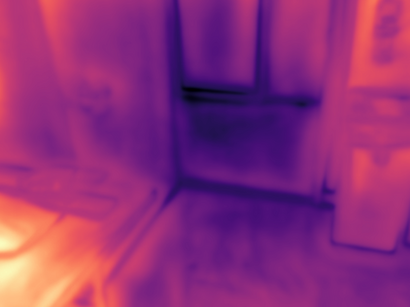} &
	\includegraphics[width=0.65\linewidth]{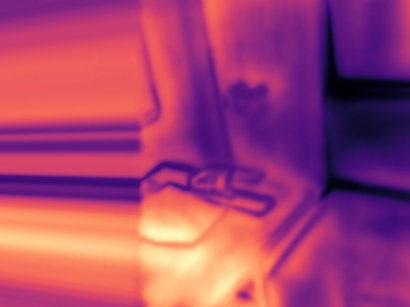} &
	\includegraphics[width=0.65\linewidth]{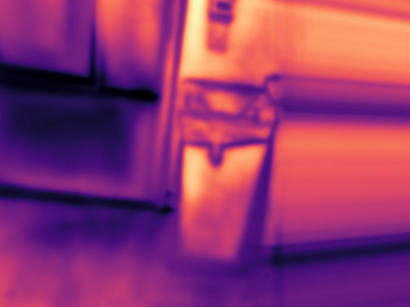} &
	\includegraphics[width=0.65\linewidth]{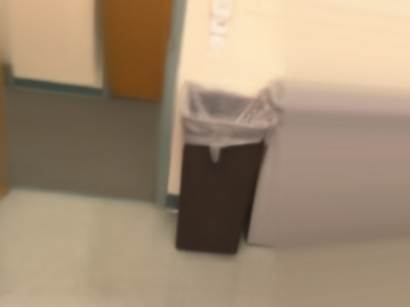} \\
	& VisionNeRF & \includegraphics[width=\linewidth]{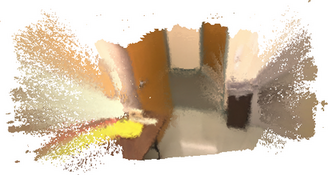} &
	\includegraphics[width=0.65\linewidth]{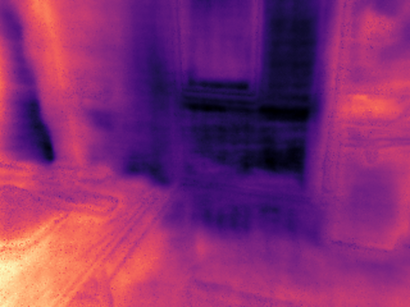} &
	\includegraphics[width=0.65\linewidth]{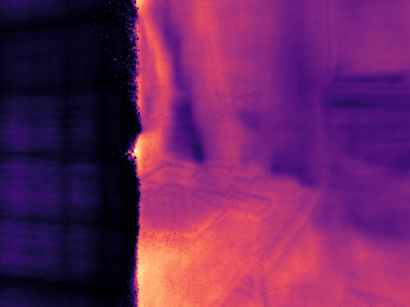} &
	\includegraphics[width=0.65\linewidth]{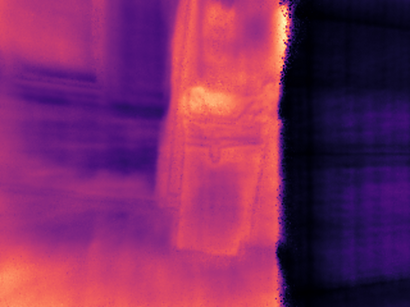} &
	\includegraphics[width=0.65\linewidth]{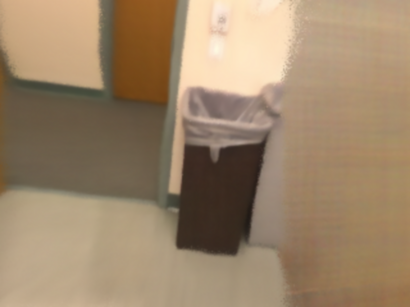} \\
	 & \best{SceneRF} &  
	\includegraphics[width=\linewidth]{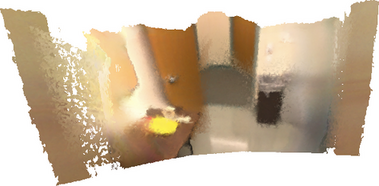} &
	\includegraphics[width=0.65\linewidth]{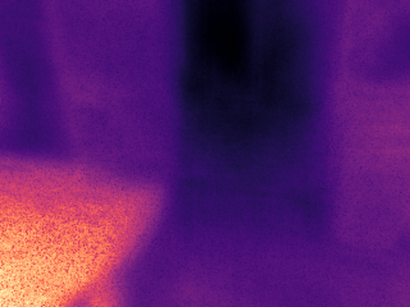} &
	\includegraphics[width=0.65\linewidth]{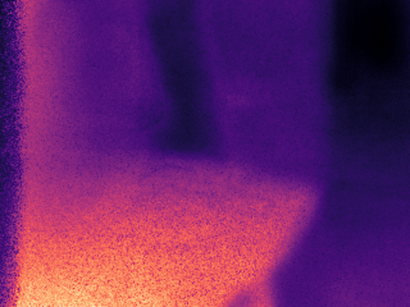} &
	\includegraphics[width=0.65\linewidth]{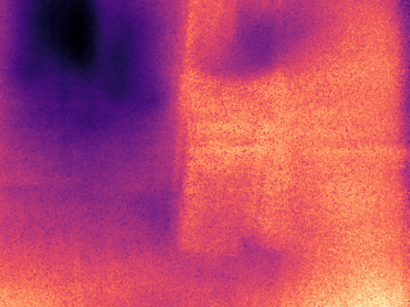} &
	\includegraphics[width=0.65\linewidth]{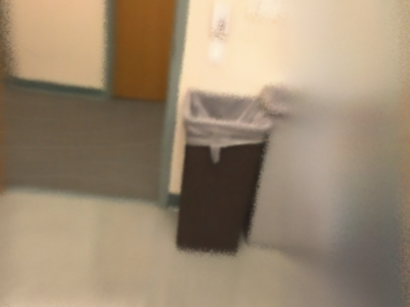} \\[-0.5em]
	\arrayrulecolor{gray!50}
	
\end{tabular}		
\caption{\textbf{Qualitative results on SemanticKITTI and BundleFusion.}
For each row, we report novel depths/views at varying positions and viewing angles w.r.t. the input frame. We note that our depths are sharper and better at-far distances. %
To produce 3D meshes, all --- even baselines --- use our scheme for reconstruction (\cref{sec:reconstruction}). 
On both datasets, our reconstruction is evidently better than others.
Please zoom in and refer to video in supplementary for better qualitative judgement.
}
\label{fig:qualitative_kitti}
\end{figure*}

\cref{fig:qualitative_kitti} also shows reconstructed 3D meshes for sample inputs. Results are better seen when zooming in and in supplementary video. On both datasets SceneRF produces better reconstruction results with less artefacts, especially on vegetation and sidewalk on SemanticKITTI and general scene structure on BundleFusion.

\subsection{Ablation studies}
\label{sec:exp_ablation}
\paragraph*{Architectural components.} 
\cref{tab:arch_ablation} reports novel depth/view synthesis of SceneRF when removing
the rgb loss ($\mathcal{L}_\text{rgb}$), reprojection loss ($\mathcal{L}_\text{reproj}$, \cref{eq:reproj}), Spherical U-Net (SU-Net, \cref{sec:sphere}), or Probabilistic Sampling (PrSamp, \cref{sec:ray_sampling}). Without SU-Net, we use a standard U-Net of similar capacity where $\psi(.)$ is a simple cartesian projection. Without PrSamp, we {revert} to standard hierarchical sampling~\cite{nerf,pixelnerf}, using the same number of inferences for a fair comparison.

In a nutshell, all our components contribute to the best novel depth synthesis metrics. In particular, $\mathcal{L}_\text{reproj}$ and PrSamp improve significantly the absolute relative error and the $\delta{}1$, showing a beneficial effect on close range depth estimation. 
For the subsidiary task of novel view synthesis, our components have mixed effects showing that depth improvement comes {at the cost of slightly lower image reconstruction}. %

\paragraph*{Probabilistic Ray Sampling (\cref{sec:ray_sampling}).} 
It is tempting to assume that PrSamp would better approximate the underlying density volume with more Gaussians or more sampled points, thus yielding better results. This is proven wrong in \cref{tab:prsom_ablation} where we vary the number of Gaussians~($k$) and points sampled per Gaussian~($m$). The best results are with $k{=}4$ and $m{=}8$. 
We conjecture this relates to the radiance field not being able to optimize too many surfaces per ray. Fewer Gaussians also preserve computational cost, more Gaussians introduce noise with fewer points per Gaussian. 

We now compare PrSamp ($k{=}4$ and $m{=}8$) against other samplings. First, we train SceneRF$^\text{MN360}$ where PrSamp is replaced by the sampling of MipNerf360~\cite{barron2022mipnerf360}. Our SceneRF (\ie, using PrSamp) outperforms SceneRF$^\text{MN360}$ on \textit{all} metrics \textit{and} datasets, with $\delta{}1$/$\delta{}2$/$\delta{}3$ of +6/+3/+2 on SemanticKITTI and +5/+4/+2 on BundleFusion.
We conjecture that this relates to our uniform sampling~($\textcolor{green}\blacktriangle$,~\cref{sec:ray_sampling}) which encourages ray exploration, \ie fighting view ambiguity, while MN360 coarse-to-fine distillation prevents escaping from invalid minima.
Importantly, note that MN360 uses 96 inferences (64 proposal+32 NeRF) and PrSamp~only 64 (32+32).
Second, we depart from original VisionNerf in \cref{tab:novelviews} and train VisionNerf$^\text{PrSamp}$ where hierarchical sampling is replaced by our PrSamp, which proves to improve $\delta{}1$/$\delta{}2$/$\delta{}3$ by +3.9/+0.1/+0.1 on SemanticKITTI.

\paragraph*{Explicit depth optimization ($\Lreproj$).}
Besides performance in~\cref{tab:arch_ablation}, it is reasonable to question the need of explicit depth optimization as NeRF-based methods can implicitly estimate depth.
We argue that $\Lrgb$ and $\Lreproj$ pursue slightly different objectives since $\Lrgb$ optimizes the rendered image by adjusting point density color~$\mathbf{c}$ and $\sigma$ w.r.t. \textit{source frame} ($I_j$ in~\cref{fig:overview}), while $\Lreproj$ optimizes reprojection of source on target~($I_{j-1}$ in~\cref{fig:overview}) but \textit{solely by adjusting depth} with $\sigma$.
In~\cref{tab:arch_ablation} \textit{bottom} we verify the complementarity of the two losses.
First, we `Freeze $\sigma$ in $\Lrgb$' to separate both optimization objectives, which performs worse (${-}4$ on $\delta_1$).
Second, we verify that using \textit{target} in $\Lreproj$ does not provide an unfair edge by removing $\Lreproj$ and replacing $\Lrgb$ with `$\Lrgb$ on source+target' --- which also drops performance (${-}6$ on $\delta_1$).
In~\cref{sec:lreproj_baselines}, we also show that $\Lreproj$ can boost the geometric ability of other NeRFs.

\begin{table}
\scriptsize
\centering
\newcolumntype{H}{>{\setbox0=\hbox\bgroup}c<{\egroup}@{}}
\setlength{\tabcolsep}{0.015\linewidth}
{
	\begin{tabular}{cc|ccccccc HHH}
		\toprule
		$k$  & $m$ & Abs Rel$\downarrow$  & Sq Rel$\downarrow$         & RMSE$\downarrow$ & RMSE log$\downarrow$& $\delta$1$\uparrow$ & $\delta$2$\uparrow$       & $\delta$3$\uparrow$ & LPIPS$\downarrow$ & SSIM$\uparrow$ & PSNR$\uparrow$ \\
		\midrule
		1  & 32 &  0.1850 & 1.358 & 5.956 & 0.2940 & 71.38 & 88.73 & 94.51 & 0.511 &	0.461 & 16.23	 \\
		\midrule
		2 & 16 &  0.1788 & 1.327 & 5.889 & 0.2878 & 72.68 & \underline{88.90} & \underline{94.70} & 0.499 &	0.467 & 16.26\\
		\midrule
		\multirow{3}{*}{4} & 4 & 0.1845  & 1.371  & 5.878 & 0.2940 & 71.62 & 88.59 & 94.51 & 0.508 & 0.461 & 16.34 \\
		& 8 &   0.1717 & \textbf{1.309} & \textbf{5.696} & \textbf{0.2809} & \textbf{75.01} & \textbf{89.35} & \textbf{94.76} & 0.490 & 0.475 & 16.29   \\
		& 16 & \textbf{0.1664}\textbf{\textbf{}} & 1.319 & 5.980 & 0.2894 & 74.58 & 88.48 & 94.17 & \textbf{0.471} & \textbf{0.483} & \textbf{16.45} \\
		\midrule
		\multirow{2}{*}{8} & 
		4 & 0.1768 & \underline{1.311} & 5.824 & 0.2910 & 72.86 & 88.60 & 94.42 & 0.499 & 0.467 & 16.28 \\
		& 8 & \underline{0.1697} & \underline{1.311} & \underline{5.794} & \underline{0.2873} & \underline{74.59} & 88.71 & 94.34 & \underline{0.484} & \underline{0.479} & \underline{16.42} \\
		\bottomrule
	\end{tabular}
}
\caption{{\textbf{PrSamp ablation on Sem.KITTI (val).} We vary number of Gaussians ($k$) and points per Gaussian ($m$).}}\label{tab:prsom_ablation}
\end{table}

\paragraph*{Spherical U-Net (\cref{sec:sphere}).}
\cref{tab:arch_ablation} `w/o SU-Net' highlights the benefit of our SU-Net. We complement this study, by comparing planar (\ie, standard decoder) and spherical decoder of different horizontal FOV. 
We experiment with planar-80$^\circ$/planar-120$^\circ$/spherical-80$^\circ$/spherical-120$^\circ$, getting respectively 17.66/17.25/17.67/\textbf{17.17} for Abs~Rel metric (lower is better) and 73.78/74.23/73.46/\textbf{75.01} for $\delta1$ (higher is better). Larger FOV seems to always improve, but our spherical decoder reaches the best results --- presumably because it induces less projection distortion.\\
\begin{figure}
	\centering
	\resizebox{1.0\linewidth}{!}
	{
		\setlength{\tabcolsep}{0.0\linewidth}
		\begin{tabular}{ccccc}
			\includegraphics[width=4em]{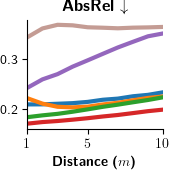}\hspace{0.2em}
			&\includegraphics[width=4em]{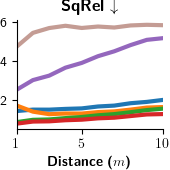}\hspace{0.2em}
			&\includegraphics[width=4em]{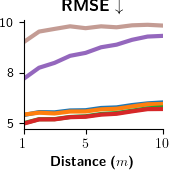}\hspace{0.2em}
			&\includegraphics[width=4em]{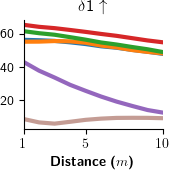}\hspace{0.2em}
			&\hspace{0.5em}\makecell{\vspace{4em}\includegraphics[width=4.5em]{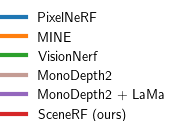}}\\[-3em]
		\end{tabular}
	}
	\caption{\textbf{Performance vs. input view distance on SemanticKITTI.} Novel depth quality drops as distance increases due to lower overlaps of FOV with the input view.}
	\label{fig:performance_fall_off}
\end{figure}

\paragraph*{Performance beyond input FOV.} 
Different than generative methods, like GAN, a minimum FOV overlaps between the input and the novel view is needed to estimate relevant features. We quantify this on novel depth in \cref{fig:performance_fall_off} showing that all metrics drop significantly as a function of the novel view distance although SceneRF is consistently better.
For novel view synthesis, we evaluate the quality of the generated \textit{unseen} pixels using `masked metrics' in~\cref{tab:masked_metrics}, \ie, evaluating only pixels \textit{not seen} in the input frame. Here again, SceneRF is far better than any other baselines.

\begin{table}
	\centering
	\setlength{\tabcolsep}{0.008\linewidth}
	\resizebox{1.0\linewidth}{!}{
		\begin{tabular}{c|c|ccccccc|ccc}
			\toprule
			& & \multicolumn{7}{c|}{\bf Novel depth synthesis} & \multicolumn{3}{c}{Novel view synthesis} \\
			& Method & AbsRel$\downarrow$ & SqRel$\downarrow$ & RMSE$\downarrow$ & RMSElog$\downarrow$ & $\delta$1$\uparrow$ & $\delta$2$\uparrow$ & $\delta$3$\uparrow$ & LPIPS$\downarrow$ & SSIM$\uparrow$ & PSNR$\uparrow$ \\
			\midrule 
			\multirow{4}{*}{\rotatebox[origin=c]{90}{SemKITTI}} & PixelNeRF & 0.5145 & 8.057 & 14.835 & 0.843 & 9.10 & 26.68 & 47.58    & \multicolumn{2}{c}{\multirow{4}{*}{\makecell{N / A}}} & 33.48  \\
			& MINE & 0.3869 & 6.099 & 13.105 & 0.656 & 25.41 & 50.43 & 67.55  &&& 33.47 \\
			& VisionNerf & 0.4831 & 7.556 & 14.573 & 0.825 & 14.50 & 34.43 & 52.44  &&& 33.41 \\
			& SceneRF & \best{0.3056} & \best{4.187} & \best{9.980} & \best{0.447} & \best{44.32} & \best{69.56} & \best{81.40}  &&& \best{33.91} \\
			\midrule
			\multirow{4}{*}{\rotatebox[origin=c]{90}{Bun.Fusion}} & PixelNeRF & 3.2717 & 20.369 & 5.277 & 1.441 & 4.48 & 10.40 & 15.75 & \multicolumn{2}{c}{\multirow{4}{*}{\makecell{N / A}}} & 22.18 \\
			& MINE & 0.2047 & 0.112 & 0.388 & 0.246 & 62.77 & 90.90 & 98.24 &&&  25.47 \\
			& VisionNerf & 3.3925 & 20.645 & 5.360 & 1.453 & 4.43 & 10.11 & 14.67 &&& 21.63 \\
			& SceneRF & \best{0.1848} & \best{0.092} & \best{0.343} & \best{0.211} & \best{70.06} & \best{94.00} & \best{99.18} &&& \best{25.90} \\
			\bottomrule
		\end{tabular}
	}
	\caption{\textbf{Masked metrics.} We calculate the metrics for pixels that are \textit{not visible} in the input image, highlighting the superiority of SceneRF compared to the baselines.}
	\label{tab:masked_metrics}
\end{table}

\begin{table}
\scriptsize
\centering
\setlength{\tabcolsep}{0.015\linewidth}
{
	\begin{tabular}{l|c|ccc|ccc}
		\toprule
		\multicolumn{2}{c}{} & \multicolumn{3}{c}{\textbf{SemanticKITTI}}& \multicolumn{3}{c}{\textbf{BundleFusion}}\\
		Method & \makecell{Need\\depth} & IoU & Prec. & Rec. & IoU & Prec. & Rec. \\
		\midrule
		AdaBins~\cite{adabin} & \cmark & 15.37 & 27.33 & 26.00  & 18.37 & 20.65 &	62.39 \\
		\midrule
		Monodepth2*~\cite{monodepth2} & \multirow{7}{*}{\xmark} &  10.76 & 18.28 & 20.74  & 14.52 & 20.14 & 34.29 \\
		SynSin~\cite{synsin}  &  & 7.84 &  13.05 & 16.43 & 9.81 & 16.62 & 19.30 \\
		MINE~\cite{mine2021}  &  & 10.93 &  18.44 & 21.20 & 12.61 & 18.46 & 28.46 \\
		VisionNeRF~\cite{visionnerf}  &  & 11.77 & 20.14 & 22.08 & 13.65 & 20.19 & 29.65 \\
		PixelNeRF~\cite{pixelnerf}  &  &  11.65 & 19.73 & 22.16 & 13.48 & 19.78 & 29.75 \\
		SceneRF (w/o Scheme) & & 11.80 & \best{19.91} & 22.47 & 17.33  & 20.13 &	\textbf{55.43} \\
		SceneRF & & \best{13.84} & 17.28 & \best{40.96} & \textbf{20.16} & \textbf{25.82} & 47.92 \\
		\bottomrule
	\end{tabular}\\
	{\scriptsize * Monodepth2 is trained with GT poses for fair comparison with our setting.}
}
\caption{\textbf{Variations of scene reconstruction.} We compare SceneRF against reconstruction with AdaBins~\cite{adabin} (depth-supervised) or Monodepth2~\cite{monodepth2} (self-supervised), and also report result w/o our Reconstruction Scheme (\cref{sec:reconstruction}). Note that, conversely to SceneRF, baselines use TSDF of the depth from the input view.
}%
\label{tab:scenereconsabl}
\end{table}

\paragraph*{Scene reconstruction (\cref{sec:reconstruction}).} 
We study variations of our scene reconstruction scheme in~\cref{tab:scenereconsabl}.
In the first 3 rows, we evaluate reconstruction using \textit{a single depth map at the input frame} with the best monocular depth estimation methods being: AdaBins~\cite{adabin} (depth-supervised), Monodepth2~\cite{monodepth2}, and SceneRF w/o reconstruction scheme. AdaBins is the only that requires depth and logically outperforms others on SemKITTI where scene are deep and Lidar provides an unfair supervision edge. On~\makebox{BundleFusion}, SceneRF however outperforms AdaBins by \makebox{${\approx{}}+2$~IoU} points which is remarkable as it is self-supervised.
SceneRF also reaches best performance among all self-supervised methods with roughly +3 and +6 IoU w.r.t. Monodepth2~\cite{monodepth2} on SemKITTI and BundleFusion, respectively. SceneRF also outperforms reconstructions from SynSin and all other NeRFs by a few points on both datasets. In~\cref{sec:supp:ablations}, we also study the effect of varying steps ($\rho$) and rotations ($\Phi$) in our reconstruction scheme.

\section{Discussion}
\label{sec:discussion}

{To the best of our knowledge, SceneRF is the first method to handle complex cluttered scenes. Still, self-supervised monocular scene reconstruction is yet in its early steps, and we discuss here some remaining challenges.} 

\begin{figure}
	\centering
	\newcolumntype{P}[1]{>{\centering\arraybackslash}m{#1}}
	\setlength{\tabcolsep}{0.004\textwidth}
	\renewcommand{\arraystretch}{0.8}
	\resizebox{1.0\linewidth}{!}{%
		\tiny
		\begin{tabular}{P{0.15\linewidth}  P{0.2\linewidth}  P{0.18\linewidth} P{0.18\linewidth} P{0.18\linewidth}}		
			\multirow{2}{*}{Input}   & \multirow{2}{*}{3D mesh}  &  \multicolumn{2}{c}{Novel depth}     &  {Novel view} \\
			& &  $+1$m, $0^{\circ}$ & $+3$m,${-}10^{\circ}$ & ${+}3$m,${-}10^{\circ}$ \\
			
			\includegraphics[width=\linewidth]{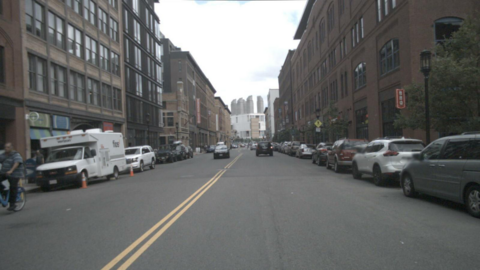} &  
			\includegraphics[width=\linewidth]{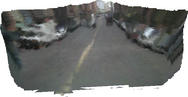} &
			\includegraphics[width=\linewidth]{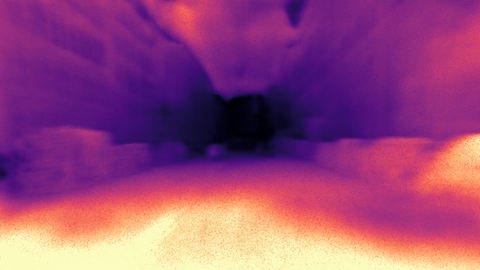} &
			\includegraphics[width=\linewidth]{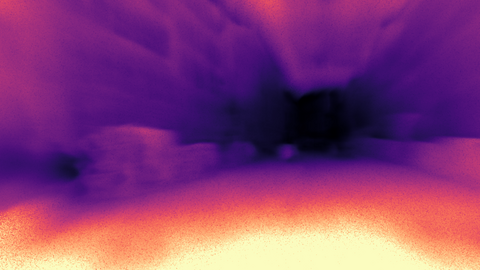} &
			\includegraphics[width=\linewidth]{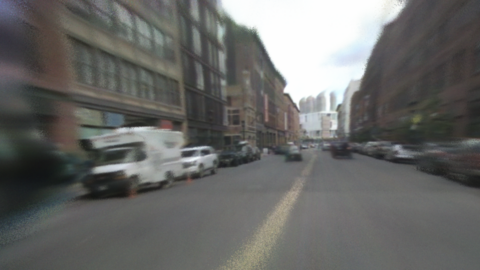} \\

			\includegraphics[width=\linewidth]{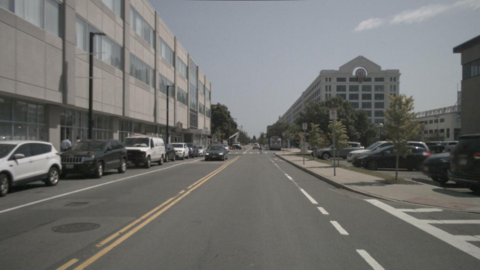} &  
			\includegraphics[width=\linewidth]{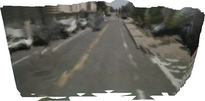} &
			\includegraphics[width=\linewidth]{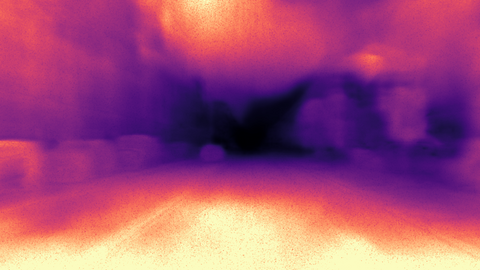} &
			\includegraphics[width=\linewidth]{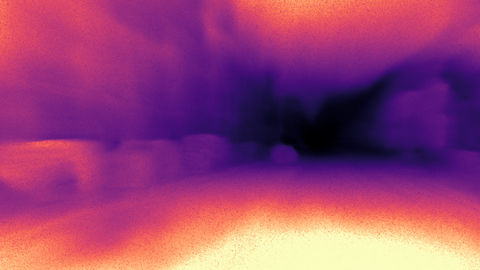} &
			\includegraphics[width=\linewidth]{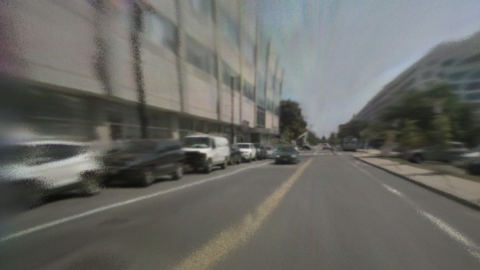} \\[-1.5em]
	\end{tabular}}
	\caption{\textbf{nuScenes generalization.} (train on SemKITTI)}
	\label{fig:reb_nuscenes}
\end{figure}

\condenseparagraph{Features compression.}
A drawback of our planar${\mapsto}$spherical mapping of SU-Net is that it induces spatial compression. An intuitive example is when input/output are of same size, since features will project on a smaller spatial portion of the output feature map. A simple workaround would be to increase output size but this would come at higher memory~cost.

\condenseparagraph{Inference time.} 
{Despite fewer inferences thanks to our PrSamp, depth synthesis is still time-consuming due to per-point inference --- which limits applicability. We conjecture that ray inference~\cite{sitzmann2021lfns} could be beneficial here.}

\condenseparagraph{Generalization.} 
To overcome the highly ill-posed problem of reconstruction from a single image, NeRF-based methods rely on strong priors learned on the training set. 
This poses inevitable issues for \textit{across domains} generalization (\eg, beyond driving scenes). 
Still, in \cref{fig:reb_nuscenes}  we show that when training on SemanticKITTI, SceneRF exhibits some generalization capability to the unseen nuScenes images~\cite{Caesar2020nuScenesAM} despite a large gap (Germany$\mapsto$USA, different \makebox{camera setup}, etc.).

\condenseparagraph{Direct Field Reconstruction.}
{As SceneRF uses fused synthesized depths (\cref{sec:reconstruction}) which are proxies of the radiance field, this suggests that reconstruction could be achieved directly. While our experiments show that using alpha/sigma to reconstruct 3D scene is not straightforward, we believe an interesting avenue for research is to seek direct extraction of surfaces from the radiance field.}\\

{\footnotesize\condenseparagraph{Acknowledgment} {The work was partly funded by the French project SIGHT (ANR-20-CE23-0016) and conducted in the SAMBA collaborative project, co-funded by BpiFrance in the Investissement d’Avenir Program. It was performed using HPC resources from GENCI–IDRIS (Grant 2021-AD011012808, 2022-AD011012808R1, and 2023-AD011014102). We thank Fabio Pizzati and Ivan Lopes for their kind proofreading and all Astra-vision group members of Inria Paris for the insightful discussions.}}

\newpage
\appendix

We begin by discussing the broader impact and ethics of our work. We then present additional ablations in~\cref{sec:supp:ablations} along with further implementation details in~\cref{sec:additional_implementation_details}. After that, we demonstrate in~\cref{sec:lreproj_baselines} that $\Lreproj$ improves all baselines. Details of the baselines are presented in~\cref{supp:sec:baselines_details}. Finally, we show additional qualitative results in~\cref{supp:sec:additional_qual_results}. 
\textbf{The supplementary video}, which is available at \href{https://astra-vision.github.io/SceneRF/}{https://astra-vision.github.io/SceneRF/}, allows better evaluation of our method.

\paragraph*{Broader impact, Ethics.} 
The promotion of self-supervised monocular 3D reconstruction contributes to alleviating the needs of costly data acquisition and labeling campaigns. 
On the long term, this also paves the way to 3D algorithms training directly on video sequences -- easier to collect and significantly more diverse than existing 3D datasets. A by-product is that it would contribute to improving generalization of 3D reconstruction.
While there are no ethical concerns specific to our proposed method, we note that all methods estimating 3D from 2D are far less precise than those leveraging depth sensors (\eg, lidar, depth cameras, stereo, etc.). When it comes to safety-critical applications, like autonomous driving, we argue for use of redundant sensors.  

\section{Additional ablations}
\label{sec:supp:ablations}

\paragraph*{Sampling strategy.}
In \cref{tab:scenereconsabl} we illustrate the effect of varying step ($\rho$) and angle ($\phi$) when sampling novel depths/views in our scene reconstruction scheme (\cref{sec:reconstruction}). As noted in~\cref{sec:experiments}, the views are sampled up to a distance of 10m on SemanticKITTI~\cite{semkitti} and 2m on BundleFusion~\cite{dai2017bundlefusion}. 
Increasing the number of synthesized depth maps (i.e., reducing the step $\rho$) and varying the angle ($\phi$) generally improves the IoU score. However, excessively large angle ($\geq$ 20$^\circ$ on SemanticKITTI~\cite{semkitti}, $\geq$ 30$^\circ$ on BundleFusion~\cite{semkitti}) tends to degrade performance since the synthesized angle diverges significantly from the available angles during training. This is particularly pertinent in the context of autonomous driving setups with front-facing cameras, where there is limited peripheral supervision.

\begin{table}[t!]
	\scriptsize
	\centering
	\begin{subfigure}[b]{0.47\linewidth}
		\centering
		\setlength{\tabcolsep}{0.015\linewidth}
		\begin{tabular}{cc|ccc}
			\toprule
			step (m) & rot. (deg.) & IoU & Prec. & Rec. \\
			\midrule
			\multicolumn{2}{c|}{\textit{w/o sampling}} & 11.80 & \best{19.91} & 22.47 \\
			0.25 & -10 / 0 / +10 & \second{13.73} & 16.98 & \best{41.78} \\
			\rowcolor{black!10}
			0.5 & -10 / 0 / +10 & \best{13.84} & 17.28 & \second{40.96} \\
			1.0 & 0 & 13.08 & \second{18.56} & 30.68 \\
			1.0 & -10 / 0 / +10 & 13.40 & 17.27 & 37.43 \\
			1.0 & -20 / 0 / +20 & 13.37 & 16.73 & 39.97 \\
			1.0 & -30 / 0 / +30 & 13.24 & 16.40 & 40.73 \\
			2.0 & -10 / 0 / +10 & 13.35 & 17.41 & 36.35 \\
			\bottomrule
		\end{tabular}
		\caption{SemanticKITTI~\cite{semkitti}}
	\end{subfigure}
	\hfill
	\begin{subfigure}[b]{0.47\linewidth}
		\centering
		\setlength{\tabcolsep}{0.015\linewidth}
		\begin{tabular}{cc|ccc}
			\toprule
			step (m) & rot. (deg.) & IoU & Prec. & Rec. \\
			\midrule
			\multicolumn{2}{c|}{\textit{w/o sampling}} & 17.33 &	20.13 & \textbf{55.43} \\
			0.1 & -20 / 0 / +20 & \best{20.34} & 25.70 & \second{49.28} \\
			0.2 & 0 & 18.76 & \best{26.43} & 39.28 \\ 
			0.2 & -10 / 0 / +10 & 19.94 & 25.80 & 46.76 \\
			\rowcolor{black!10}
			0.2 & -20 / 0 / +20 & \second{20.16} &	25.82 &	{47.92} \\
			0.2 & -30 / 0 / +30 & 19.98 & \second{25.90} & 46.64 \\
			0.3 & -20 / 0 / +20 & 19.84 & 25.84 &  46.09 \\
			\bottomrule
		\end{tabular}
		\caption{BundleFusion~\cite{dai2017bundlefusion}}
	\end{subfigure}
	\caption{\textbf{Sampling for reconstruction.} Performance of our reconstruction scheme when varying our sampling steps ($\rho$) and angles ($\phi$) (val. set). The highlighted row is our main setup.}
	\label{tab:scenereconsabl}
\end{table}

\paragraph*{BundleFusion evaluation on more sequences.} For completeness, we retrain on 6 sequences (apt0, apt1, apt2, office1, office2, office3) with 2 validation sequences of unseen rooms (copyroom, office0) to show generalization. The results are shown in~\cref{tab:bundlefusion_eval2} and show that SceneRF still surpasses all baselines on all 10 metrics except LPIPS. This is on par with~\cref{tab:novelviews}.
\begin{table}
	\centering
	\setlength{\tabcolsep}{0.008\linewidth}
	\resizebox{1.0\linewidth}{!}
	{
		
		\begin{tabular}{c|ccccccc | ccc}
			\toprule
			& \multicolumn{7}{c|}{\bf Novel depth synthesis} & \multicolumn{3}{c}{Novel view synthesis}\\
			Method & Abs Rel$\downarrow$  & Sq Rel$\downarrow$         & RMSE$\downarrow$ & RMSE log$\downarrow$ & $\delta$1$\uparrow$ & $\delta$2$\uparrow$       & $\delta$3$\uparrow$ & LPIPS$\downarrow$ & SSIM$\uparrow$ & PSNR$\uparrow$ \\
			\midrule 
			PixelNeRF   &  0.6332  &  2.370 & 1.786 & 0.5722 & 48.01 & 73.82  & 84.09 &   0.421  & 0.770 & 19.36 \\
			MINE     &  \second{0.1712} &  \second{0.085} & \second{0.368}  & \second{0.2214} &  \second{70.74}   &  \second{93.42} & \second{98.46} & 0.430 &	0.714 & \second{20.21}\\
			VisionNerf &  0.6384 & 2.813  & 1.934 &  0.5721 & 59.25  & 78.62  & 84.37  & \best{0.391} & 	\second{0.790} & 19.67	 \\
			SceneRF  & \best{0.1581}	& \best{0.069}  &  \best{0.330} & \best{0.1921} & \best{75.81}  & \best{96.80}  & \best{99.66} & \second{0.404} & \best{0.801} & \best{24.02}	\\
			\bottomrule
		\end{tabular}
		
	}\caption{\textbf{BundleFusion.} Train on 6 scenes and evaluate on 2 scenes.}
	\label{tab:bundlefusion_eval2}
\end{table}

\begin{table*}
	\scriptsize
	\centering
	{
		\begin{tabular}{c|c|ccccccc | ccc}
			\toprule
			& & \multicolumn{7}{c|}{Novel depth synthesis} & \multicolumn{3}{c}{Novel view synthesis}\\
			Method & $\Lreproj$ & Abs Rel$\downarrow$  & Sq Rel$\downarrow$         & RMSE$\downarrow$ & RMSE log$\downarrow$ & $\delta$1$\uparrow$ & $\delta$2$\uparrow$       & $\delta$3$\uparrow$ & LPIPS$\downarrow$ & SSIM$\uparrow$ & PSNR$\uparrow$ \\
			\midrule 
			\multirow{2}{*}{PixelNeRF~\cite{pixelnerf}}  & \xmark &   0.2364 & 2.080 & 6.449 & 0.3354 & 65.81 & 85.43 & 92.90 & 0.489 & 0.466 & 15.80      \\
			& \checkmark  & \best{0.1986} & \best{1.544} & \best{5.963} & \best{0.3093} & \best{70.30} & \best{87.19} & \best{93.82} &  \best{0.488} & \best{0.481} & \best{16.11}     \\
			\midrule 
			
			\multirow{2}{*}{MINE~\cite{mine2021}}  & \xmark &    0.2248 & 1.787 & 6.343 & 0.3283 & 65.87	& 85.52 & 93.30 &  0.448  &  0.496 &     \best{16.03}       \\
			& \checkmark & \best{0.2003} & \best{1.599} & \best{6.023} & \best{0.3070} & \best{70.22} & \best{86.98} & \best{93.89} & \best{0.445} & \best{0.497} &  15.96 \\
			
			\midrule 
			\multirow{2}{*}{VisionNerf~\cite{visionnerf}} & \xmark & 0.2054 & 1.490 & 5.841 & 0.3073 & 69.11 & 88.28 & 94.37 & 0.468 & 0.483 & \best{16.49} \\			
			& \checkmark & \best{0.1749} & \best{1.380} & \best{5.643} & \best{0.2841} & \best{75.77} & \best{89.25} & \best{94.58} & \best{0.432} & \best{0.488} &  16.39\\
			
			\bottomrule
		\end{tabular}
	}
	\vspace{1pt}
	
	\caption{\textbf{Reprojection loss $\Lreproj$ on other baselines.} We apply our reprojection loss to other NeRF baselines, showing it boosts performance for all.}\label{tab:reproject_loss_baselines}
\end{table*}

\section{Effect of \texorpdfstring{$\Lreproj$}{TEXT} on NeRF baselines}
\label{sec:lreproj_baselines}
In \cref{tab:reproject_loss_baselines}, we apply the reprojection loss $\Lreproj$ (\cref{sec:depth_optim}) to all NeRF baselines, showing that it improves consistently all of them. Again, we argue this is because $\Lreproj$ enforces better density ($\sigma$) in the volume rendering -- which has a complementary effect with $\Lrgb$.

\section{Additional implementation details}
\label{sec:additional_implementation_details}

\subsection{Probabilistic ray sampling (PrSamp) details}
\label{sec:prsamp_detail}
For clarity, in \cref{algo:prsamp}, we detail the pseudocode of the Probabilistic Ray Sampling (\cref{sec:ray_sampling}).

\begin{algorithm*}[t]
	\footnotesize
	\SetAlgoLined
	\SetKwFunction{Gsample}{gauss-sampling}
	\SetKwFunction{Uniform}{uniform-samp}
	\SetKwFunction{PrSOM}{PrSOM}
	\SetKwFunction{KL}{Kullback-Leibler}
	\SetKwFunction{Alpha}{alpha-value}
	\SetKwFunction{Train}{train}
	\SetKwFunction{Direction}{dir}
	\SetKwFunction{GD}{GD}
	\SetKwInOut{Input}{Input}  
	\SetKwInOut{Parameter}{Param}
	\Input{Ray $\mb{r}$. \\
	}
	\Parameter{
		Number of Gaussians $k$, and $m$ number of points per Gaussian.
		\\
		Near and far bounds: $t_\text{n}=0.2\text{m}$ and $t_\text{f}=100\text{m}$.
		\\
		Learning rate $lr$ of gradient descend (GD).\\
	}
	\KwResult{Points sampled $\mc{P}$} 
	$\mb{d} \gets \Direction(\mb{r})$ \\
	\tcp{Uniform sampling ($\textcolor{blue}\bullet$)}
	$\mc{I} \gets \{\Uniform(\text{num=}k, \text{start=}t_\text{n}, \text{end=}t_\text{f})\times\mb{d}\}$ \Comment{Points sampling between near and far bounds}\\
	\tcp{\step{1} Predicts Gaussians ($\mc{G}$) with MLP $g(\cdot)$}
	$\mc{G} \gets g\big(\big\{(\mb{x}, \mb{W}(\psi(\mb{x})))\,|\, \forall \mb{x} \in \mc{I}\big\}\big)$ \\
	\tcp{\step{2} Sample $m$ points from Gaussians ($\textcolor{orange}\blacksquare$)}
	$\mc{P} \gets \emptyset$\\
	\For{$i \gets 1$ to $k$}{%
		$\mc{P} \gets \mc{P} \cup \Gsample(\mc{G}_i, m)$ \\
	}
	\tcp{Sample 32 points uniformly ($\textcolor{green}\blacktriangle$)}
	$\mc{P} \gets \mc{P} \cup \{\Uniform(\text{num=}32, \text{start=}t_\text{n}, \text{end=}t_\text{f})\times\mb{d}\}$\\
	\tcp{\step{3}-\step{4} NeRF inference to compute densities}
	$\sigma \gets \{f(\gamma(\mb{x}),\mb{d};\mb{W}(\psi(\mb{x})))_\sigma\,|\, \forall \mb{x} \in \mc{P}\}$ \Comment{Densities from $f(\cdot)$ inferences~\cref{eq:f}} \\
	\tcp{\step{5} PrSOM point-Gaussian assigment}
	$\alpha \gets \{\Alpha(s, \dots) \,|\, \forall s \in \sigma\}$ \Comment{Compute alpha values from \cite{mildenhall2019llff} p3} \\
	$\mc{X} \gets \{\PrSOM(\mc{G}, \mc{P}, \alpha)\}$ \Comment{Applies PrSOM \cite{prsom}} \\
	\tcp{\step{6} Compute new Gaussians from assigned points and update $g(\cdot)$}
	$\mc{G}' \gets \{(\mu(\mc{X}_i), \text{std}(\mc{X}_i)) \,|\, \forall i \in \mathbb{N}, 1<=i<=k\}$ \\
	$\mc{L}_\text{gauss} \gets \frac{1}{\ngauss}\sum_i^k \KL(\gauss_i||\gauss'_i)$ \\
	$\mc{L}_\text{surface} \gets \min_i (||\mu(\gauss'_i) - \Dest||_1)$ \\
	$\mc{L}_\text{total} \gets \mc{L}_\text{gauss}+\mc{L}_\text{surface}$ \\
	$g \gets \GD^{lr}(g, \nabla\mc{L}_\text{total})$ \Comment{Applies gradient-descent to update $g(\cdot)$}
	\caption{Probabilistic Ray Sampling.}
	\label{algo:prsamp}
\end{algorithm*}

\subsection{3D reconstruction details}
\label{supp:sec:recon_detail}

\paragraph{Fusing TSDFs.} From~\cref{sec:reconstruction}, we fuse individual TSDFs by taking the minimum of their absolute values (`min') instead of the more standard average of all TSDFs (`avg'). We justify this choice, in \cref{tab:scenerecons_tsdf} showing that using `min' leads to +2.77 IoU. We argue that some surfaces may be better estimated from specific camera locations. Averaging all (`avg') has a smoothing effect on $V(\cdot)$ which subsequently reduces accuracy.

\begin{table}
	\scriptsize
	\centering
	\setlength{\tabcolsep}{0.011\linewidth}
	{
		\begin{tabular}{c|ccc}
			\toprule
			Method  & IoU & Prec. & Rec. \\
			\midrule 
			SceneRF (avg)  & 11.07 & 11.81 & 63.98 \\
			SceneRF (min)  & 13.84 & 17.28 & 40.96\\
			\bottomrule
		\end{tabular}
	}
	\caption{\textbf{TSDF fusion strategy comparison on SemanticKITTI~\cite{semkitti}}. We show that our way of extracting the TSDF described in section 3.4 is better than the traditional way of using the weighted average of TSDFs.}
	\label{tab:scenerecons_tsdf}
\end{table}

\begin{figure}
	\centering
	\footnotesize
	\includegraphics[width=0.6\linewidth]{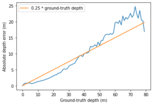}
	\caption{\textbf{Absolute depth error w.r.t. ground-truth depth.} We compute error from 100 randomly selected scenes in the training set, and observe a linear relation between error and distance.}
	\label{fig:depth_error}
\end{figure}

\paragraph{Occupancy grid.} 
To convert the scene TSDF volume $V(\cdot)$ (cf.~\cref{sec:reconstruction}) into an occupancy grid, we first study the depth estimation error.
Comparing 100 frames in \cref{fig:depth_error} with sparse Lidar ground truth, we note a linear relation between error estimation and ground truth depth. This motivated us to model the occupancy grid $O(\cdot)$ as an adaptive depth threshold:
\begin{equation}
	O(v) = 1 \iff \text{V}(v) < \min(0.25 d_v, 4.0)\,,
\end{equation}
with $v$ a voxel in $\text{V}$, and $d_v$ its distance to the camera origin. We arbitrarily cap the threshold to 4 meters to avoid considering all far voxels as occupied.

\subsection{Network architecture}
\label{supp:sec:network_architecture}
For 2D features extraction, the encoder is similar to~\cite{monoscene}, which is based on a pre-trained EfficientNetB7~\cite{effcientnet}. The spherical decoder has 5 layers, each of which doubles the input resolution and halves the feature dimension. To make up for the large amount of empty space that comes with increasing the field of view, we augment the receptive field by putting three ResNet blocks with dilation sizes of 1, 2, and 3 in each layer. 
The skip connections (described in~\cref{sec:sphere}) are used between the encoder and decoder at the corresponding scale.

\subsection{BundleFusion -- 3D ground-truth}
We seek to generate for each input frame the 3D ground-truth occupancy grid. To do this, we use the camera parameters and define a volume of (4.8m, 4.8m, 3.84m) in front of the camera. The origin is set at (-2.4m, -2.4m, 0m) such that the camera is in the middle of one side of the volume and pointing inward. 
We then combine the depth maps to create the TSDF volume with a voxel size of 0.04m, resulting in a TSDF grid of size (120, 120, 96). Finally, the occupancy grid is obtained by thresholding the TSDF grid.

\section{Baselines details}
\label{supp:sec:baselines_details}
We re-train all baseline networks, including the novel depth/view synthesis~(\cref{supp:sec:novel_depth_view_baselines}) and scene reconstruction baselines~(\cref{supp:sec:reconstruction_baselines}). We provide the reader with additional details about our baselines.

\subsection{Novel depth/views baselines}
\label{supp:sec:novel_depth_view_baselines}
We train our method and baselines using AdamW~\cite{adamW} optimizer on 4 Tesla V100 32g with learning rate of 1e-5 for 50 epochs. 
For each baseline, we rely on the recommended learning rate scheduler and number of positional encoding frequencies. For our network, since we build on PixelNeRF~\cite{pixelnerf}, we use its scheduler and number of frequencies. The ray batch size was 1200 for Semantic KITTI~\cite{semkitti} and 2048 for BundleFusion~\cite{dai2017bundlefusion}. The training time was around 5 days per network. Additional information about the baselines implementations is provided below.

\paragraph{PixelNeRF~\cite{pixelnerf}.} We use the official implementation\footnote{\url{https://github.com/sxyu/pixel-nerf}}. Following the official sampling strategy, we sample 96 points per ray, consisting of 64 coarse points, which are used to sample 16 fine points hierarchically and 16 points around the estimated depth.

\paragraph{MINE~\cite{mine2021}.} We use the official implementation\footnote{\url{https://github.com/vincentfung13/MINE}}. To balance memory cost, we use the 32 planes version.

\paragraph{VisionNeRF~\cite{visionnerf}.} We use the official implementation\footnote{\url{https://github.com/ken2576/vision-nerf}}. To balance memory cost again, we sample 96 points (32 coarse, 64 fine) which is more than for SceneRF. 

\subsection{Scene reconstruction baselines}
\label{supp:sec:reconstruction_baselines}
Only MonoScene\footnote{\url{https://github.com/cv-rits/MonoScene}} is a monocular baseline. To better compare with the literature, we follow the recommendation of MonoScene authors~\cite{monoscene} and compare against the $^\text{rgb}$ versions of popular semantic scene completion baselines: LMSCNet\footnote{\url{https://github.com/cv-rits/LMSCNet}}~\cite{lmscnet}, 3DSketch\footnote{\url{https://github.com/charlesCXK/TorchSSC}}~\cite{3DSketch} and AICNet\footnote{\url{https://github.com/waterljwant/SSC}}~\cite{aicnet}.  
More in depth, to convert the sequence of depths into 3D label to train the scene reconstruction baselines, we use the Adabin~\cite{adabin} model to predict the depth for each image and fuse all depths into a single TSDF volume, then turned into an occupancy grid with the same reconstruction scheme as for SceneRF (see~\cref{supp:sec:recon_detail}). 
For all, the mesh is obtained with the traditional marching cubes~\cite{marchingcubes}.

\section{Additional qualitative results}
\label{supp:sec:additional_qual_results}

\begin{figure}
	\centering
	\newcolumntype{P}[1]{>{\centering\arraybackslash}m{#1}}
	\setlength{\tabcolsep}{0.004\textwidth}
	\renewcommand{\arraystretch}{0.8}
	\resizebox{1.0\linewidth}{!}{%
		\footnotesize
		\begin{tabular}{P{0.30\linewidth}  P{0.3\linewidth}  P{0.3\linewidth}}		
			Input   & MonoScene (Depth-supervised) & SceneRF (Self-supervised)  \\
			\includegraphics[width=\linewidth]{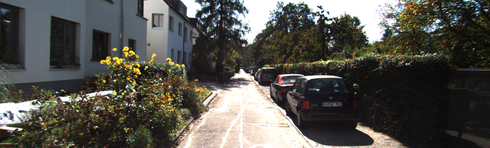} &  
			\includegraphics[width=\linewidth]{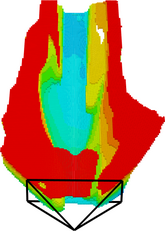} &
			\includegraphics[width=\linewidth]{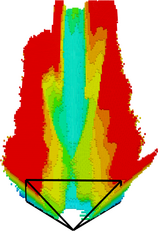} \\
			
			\includegraphics[width=\linewidth]{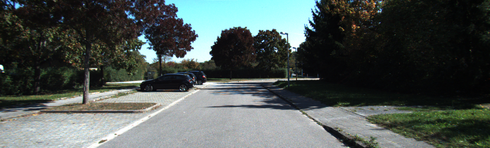} &  
			\includegraphics[width=\linewidth]{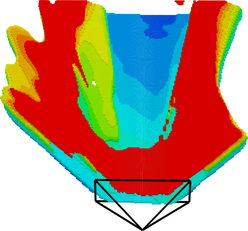} &
			\includegraphics[width=\linewidth]{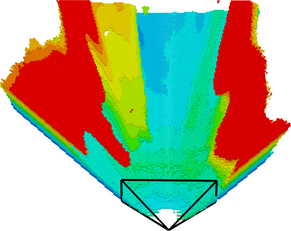} \\

			\includegraphics[width=\linewidth]{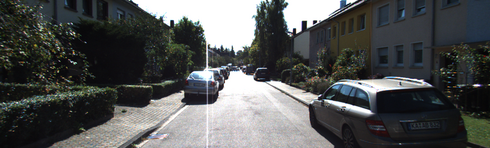} &  
			\includegraphics[width=\linewidth]{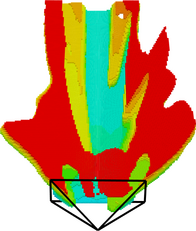} &
			\includegraphics[width=\linewidth]{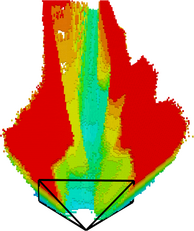} \\
			
			\includegraphics[width=\linewidth]{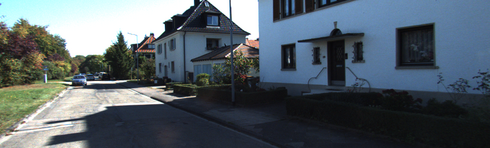} &  
			\includegraphics[width=\linewidth]{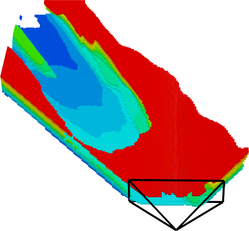} &
			\includegraphics[width=\linewidth]{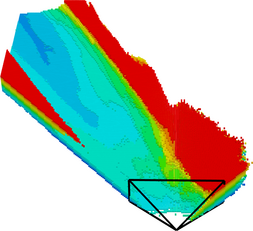} \\

			\includegraphics[width=\linewidth]{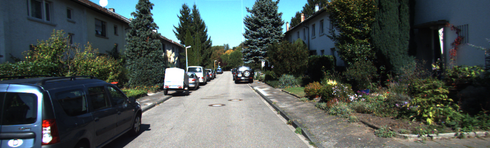} &  
			\includegraphics[width=\linewidth]{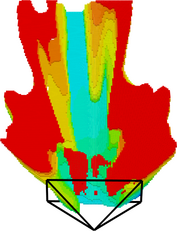} &
			\includegraphics[width=\linewidth]{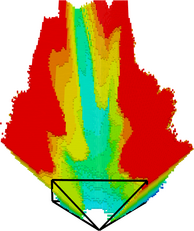} \\

			\\[-1.5em]
	\end{tabular}}
	\caption{\textbf{Voxelized reconstructions on SemanticKITTI (val set).}}
	\label{fig:voxelized_output}
\end{figure}

\paragraph{Voxelized reconstructions on SemanticKITTI~\cite{semkitti}.}
\cref{fig:voxelized_output} shows the voxelized reconstructions comparing our self-supervised SceneRF with the \textit{Depth-supervised} version of MonoScene~\cite{monoscene} which relies on AdaBins~\cite{adabin} trained with Lidar ground truth. Notably, despite less supervision the predictions of SceneRF align closely with those of MonoScene in terms of overall scene architecture, object shapes, and positioning. Intriguingly, SceneRF infers the sky more accurately, attributed to the 3D consistency gained from optimizing the radiance volume. On the other hand, MonoScene struggles to predict sky arguably because AdaBins trains with lidar depth which cannot capture the sky. Nevertheless, occlusions artefacts are visible in both due to monocular supervision.

\begin{figure*}[!t]
	\centering
	\newcolumntype{P}[1]{>{\centering\arraybackslash}m{#1}}
	\setlength{\tabcolsep}{0.004\textwidth}
	\renewcommand{\arraystretch}{0.8}
	\footnotesize
	\begin{tabular}{P{0.12\textwidth}  P{0.10\textwidth}  P{0.18\textwidth}  P{0.14\textwidth} P{0.14\textwidth} P{0.14\textwidth}  P{0.14\textwidth}}		
		\multirow{2}{*}{Input}  & \multirow{2}{*}{Method}  & \multirow{2}{*}{3D mesh}  &  \multicolumn{3}{c}{Novel depth}     &  {Novel view} \\
		\cmidrule{4-6}%
		& & & $+1$m, $0^{\circ}$ & $+3$m, ${-}10^{\circ}$ & ${+}5$m, ${+}10^{\circ}$ & ${+}5$m, ${+}10^{\circ}$ \\

		\multirow{13}{*}{\includegraphics[width=\linewidth]{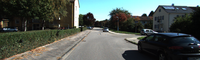}} 
		& PixelNeRF & \includegraphics[width=\linewidth]{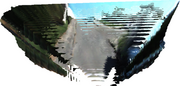} &
		\includegraphics[width=\linewidth]{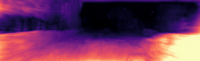} &
		\includegraphics[width=\linewidth]{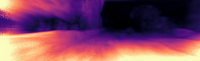} &
		\includegraphics[width=\linewidth]{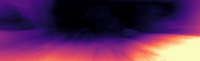} &
		\includegraphics[width=\linewidth]{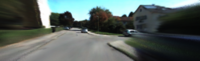} \\
		& MINE & \includegraphics[width=\linewidth]{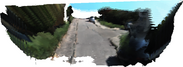} &
		\includegraphics[width=\linewidth]{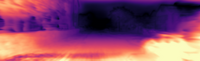} &
		\includegraphics[width=\linewidth]{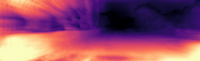} &
		\includegraphics[width=\linewidth]{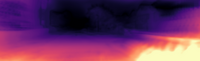} &
		\includegraphics[width=\linewidth]{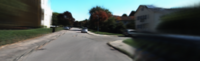} \\
		& VisionNeRF & \includegraphics[width=\linewidth]{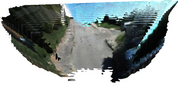} &
		\includegraphics[width=\linewidth]{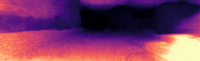} &
		\includegraphics[width=\linewidth]{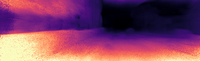} &
		\includegraphics[width=\linewidth]{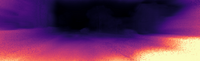} &
		\includegraphics[width=\linewidth]{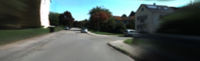} \\
		& \best{SceneRF} &  \includegraphics[width=\linewidth]{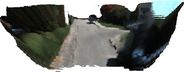} &
		\includegraphics[width=\linewidth]{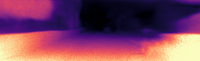} &
		\includegraphics[width=\linewidth]{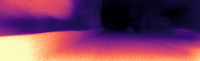} &
		\includegraphics[width=\linewidth]{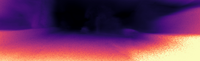} &
		\includegraphics[width=\linewidth]{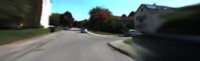} \\
		\arrayrulecolor{gray!50}
		\midrule
		\multirow{13}{*}{\includegraphics[width=\linewidth]{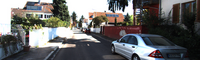}} 
		& PixelNeRF & \includegraphics[width=\linewidth]{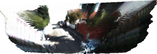} &
		\includegraphics[width=\linewidth]{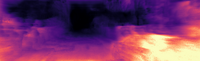} &
		\includegraphics[width=\linewidth]{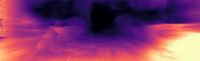} &
		\includegraphics[width=\linewidth]{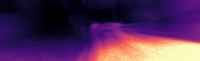} &
		\includegraphics[width=\linewidth]{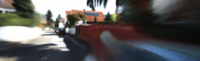} \\
		& MINE & \includegraphics[width=\linewidth]{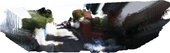} &
		\includegraphics[width=\linewidth]{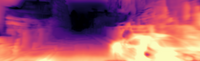} &
		\includegraphics[width=\linewidth]{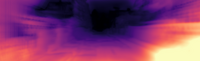} &
		\includegraphics[width=\linewidth]{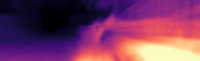} &
		\includegraphics[width=\linewidth]{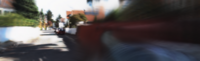} \\
		& VisionNeRF & \includegraphics[width=\linewidth]{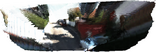} &
		\includegraphics[width=\linewidth]{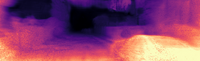} &
		\includegraphics[width=\linewidth]{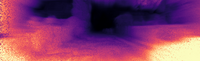} &
		\includegraphics[width=\linewidth]{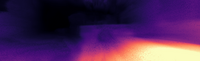} &
		\includegraphics[width=\linewidth]{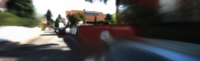} \\
		& \best{SceneRF} &  \includegraphics[width=\linewidth]{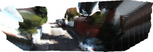} &
		\includegraphics[width=\linewidth]{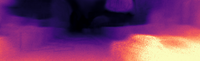} &
		\includegraphics[width=\linewidth]{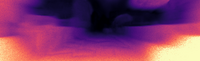} &
		\includegraphics[width=\linewidth]{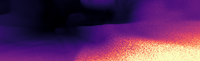} &
		\includegraphics[width=\linewidth]{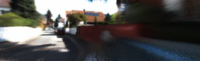} \\
		
		\arrayrulecolor{gray!50}
		\midrule
		\multirow{13}{*}{\includegraphics[width=\linewidth]{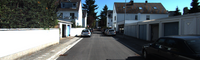}} 
		& PixelNeRF & \includegraphics[width=\linewidth]{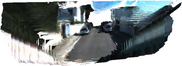} &
		\includegraphics[width=\linewidth]{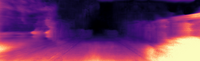} &
		\includegraphics[width=\linewidth]{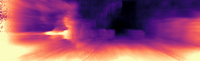} &
		\includegraphics[width=\linewidth]{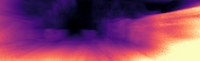} &
		\includegraphics[width=\linewidth]{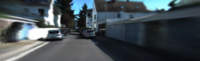} \\
		& MINE & \includegraphics[width=\linewidth]{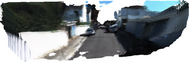} &
		\includegraphics[width=\linewidth]{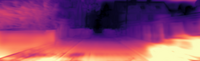} &
		\includegraphics[width=\linewidth]{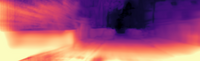} &
		\includegraphics[width=\linewidth]{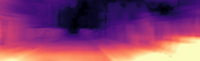} &
		\includegraphics[width=\linewidth]{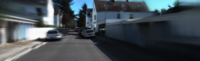} \\
		& VisionNeRF & \includegraphics[width=\linewidth]{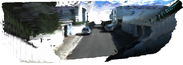} &
		\includegraphics[width=\linewidth]{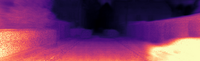} &
		\includegraphics[width=\linewidth]{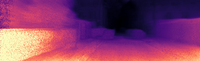} &
		\includegraphics[width=\linewidth]{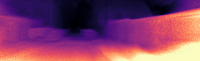} &
		\includegraphics[width=\linewidth]{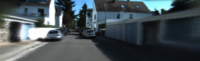} \\
		& \best{SceneRF} &  \includegraphics[width=\linewidth]{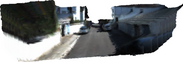} &
		\includegraphics[width=\linewidth]{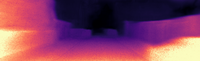} &
		\includegraphics[width=\linewidth]{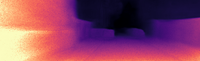} &
		\includegraphics[width=\linewidth]{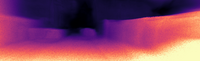} &
		\includegraphics[width=\linewidth]{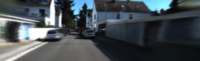} \\
		
		\arrayrulecolor{gray!50}
		\midrule
		
		\multirow{13}{*}{\includegraphics[width=\linewidth]{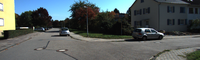}} 
		
		& PixelNeRF & \includegraphics[width=\linewidth]{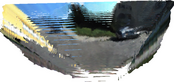} &
		\includegraphics[width=\linewidth]{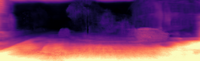} &
		\includegraphics[width=\linewidth]{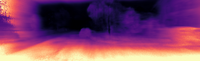} &
		\includegraphics[width=\linewidth]{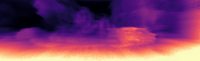} &
		\includegraphics[width=\linewidth]{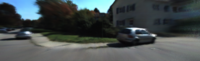} \\
		& MINE & \includegraphics[width=\linewidth]{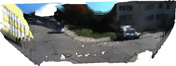} &
		\includegraphics[width=\linewidth]{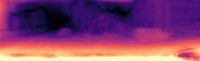} &
		\includegraphics[width=\linewidth]{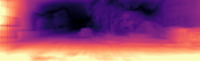} &
		\includegraphics[width=\linewidth]{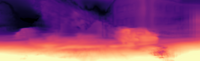} &
		\includegraphics[width=\linewidth]{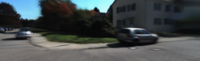} \\
		& VisionNeRF & \includegraphics[width=\linewidth]{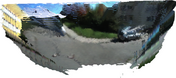} &
		\includegraphics[width=\linewidth]{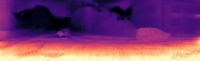} &
		\includegraphics[width=\linewidth]{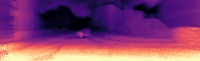} &
		\includegraphics[width=\linewidth]{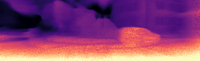} &
		\includegraphics[width=\linewidth]{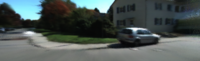} \\
		& \best{SceneRF} &  \includegraphics[width=\linewidth]{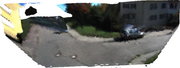} &
		\includegraphics[width=\linewidth]{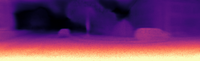} &
		\includegraphics[width=\linewidth]{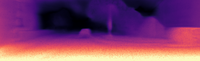} &
		\includegraphics[width=\linewidth]{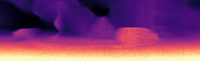} &
		\includegraphics[width=\linewidth]{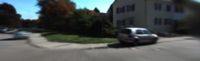} \\
		\arrayrulecolor{gray!50}
		\midrule

	\end{tabular}		
	\caption{\textbf{Additional qualitative results on SemanticKITTI~\cite{semkitti} (val.)}. }
	\label{fig:addtitonal_qualitative_kitti_1}
\end{figure*}

\begin{figure*}[!t]
	\centering
	\newcolumntype{P}[1]{>{\centering\arraybackslash}m{#1}}
	\setlength{\tabcolsep}{0.004\textwidth}
	\renewcommand{\arraystretch}{0.8}
	\footnotesize
	\begin{tabular}{P{0.12\textwidth}  P{0.10\textwidth}  P{0.18\textwidth}  P{0.14\textwidth} P{0.14\textwidth} P{0.14\textwidth}  P{0.14\textwidth}}		
		\multirow{2}{*}{Input}  & \multirow{2}{*}{Method}  & \multirow{2}{*}{3D mesh}  &  \multicolumn{3}{c}{Novel depth}     &  {Novel view} \\
		\cmidrule{4-6}%
		& & & $+1$m, $0^{\circ}$ & $+3$m, ${-}10^{\circ}$ & ${+}5$m, ${+}10^{\circ}$ & ${+}5$m, ${+}10^{\circ}$ \\
		
		\multirow{13}{*}{\includegraphics[width=\linewidth]{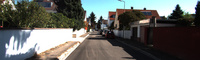}} 
		
		& PixelNeRF & \includegraphics[width=\linewidth]{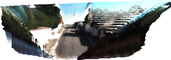} &
		\includegraphics[width=\linewidth]{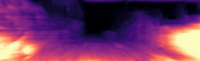} &
		\includegraphics[width=\linewidth]{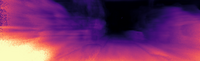} &
		\includegraphics[width=\linewidth]{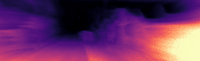} &
		\includegraphics[width=\linewidth]{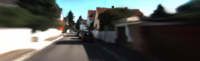} \\
		& MINE & \includegraphics[width=\linewidth]{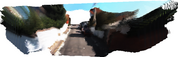} &
		\includegraphics[width=\linewidth]{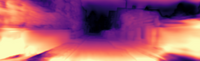} &
		\includegraphics[width=\linewidth]{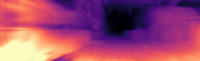} &
		\includegraphics[width=\linewidth]{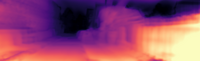} &
		\includegraphics[width=\linewidth]{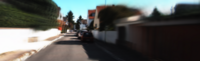} \\
		& VisionNeRF & \includegraphics[width=\linewidth]{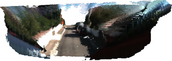} &
		\includegraphics[width=\linewidth]{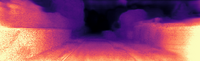} &
		\includegraphics[width=\linewidth]{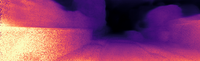} &
		\includegraphics[width=\linewidth]{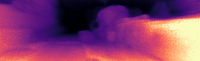} &
		\includegraphics[width=\linewidth]{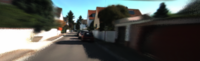} \\
		& \best{SceneRF} &  \includegraphics[width=\linewidth]{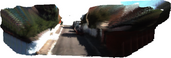} &
		\includegraphics[width=\linewidth]{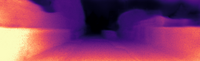} &
		\includegraphics[width=\linewidth]{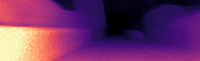} &
		\includegraphics[width=\linewidth]{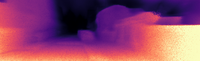} &
		\includegraphics[width=\linewidth]{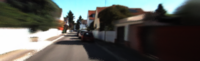} \\
		\arrayrulecolor{gray!50}
		\midrule
		
		\multirow{13}{*}{\includegraphics[width=\linewidth]{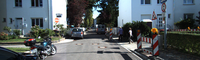}} 
		
		& PixelNeRF & \includegraphics[width=\linewidth]{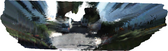} &
		\includegraphics[width=\linewidth]{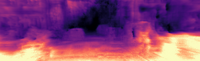} &
		\includegraphics[width=\linewidth]{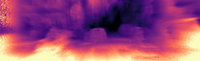} &
		\includegraphics[width=\linewidth]{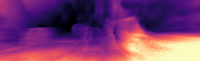} &
		\includegraphics[width=\linewidth]{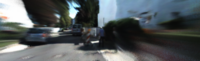} \\
		& MINE & \includegraphics[width=\linewidth]{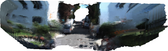} &
		\includegraphics[width=\linewidth]{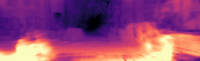} &
		\includegraphics[width=\linewidth]{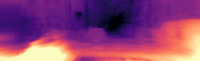} &
		\includegraphics[width=\linewidth]{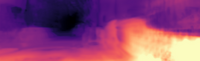} &
		\includegraphics[width=\linewidth]{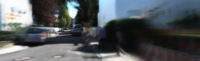} \\
		& VisionNeRF & \includegraphics[width=\linewidth]{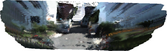} &
		\includegraphics[width=\linewidth]{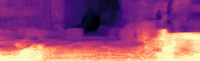} &
		\includegraphics[width=\linewidth]{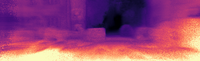} &
		\includegraphics[width=\linewidth]{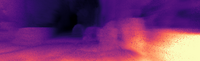} &
		\includegraphics[width=\linewidth]{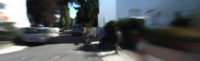} \\
		& \best{SceneRF} &  \includegraphics[width=\linewidth]{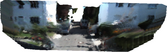} &
		\includegraphics[width=\linewidth]{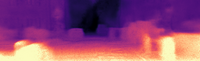} &
		\includegraphics[width=\linewidth]{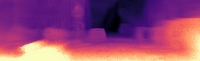} &
		\includegraphics[width=\linewidth]{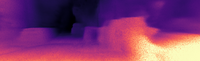} &
		\includegraphics[width=\linewidth]{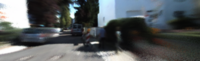} \\
		\arrayrulecolor{gray!50}
		\midrule
		
		\multirow{13}{*}{\includegraphics[width=\linewidth]{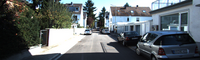}} 
		
		& PixelNeRF & \includegraphics[width=\linewidth]{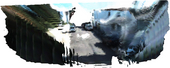} &
		\includegraphics[width=\linewidth]{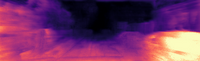} &
		\includegraphics[width=\linewidth]{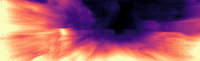} &
		\includegraphics[width=\linewidth]{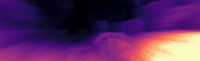} &
		\includegraphics[width=\linewidth]{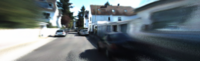} \\
		& MINE & \includegraphics[width=\linewidth]{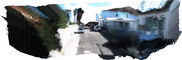} &
		\includegraphics[width=\linewidth]{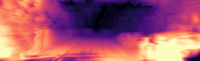} &
		\includegraphics[width=\linewidth]{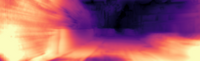} &
		\includegraphics[width=\linewidth]{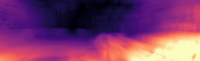} &
		\includegraphics[width=\linewidth]{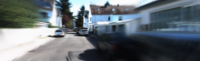} \\
		& VisionNeRF & \includegraphics[width=\linewidth]{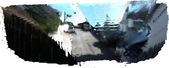} &
		\includegraphics[width=\linewidth]{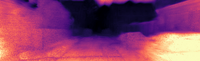} &
		\includegraphics[width=\linewidth]{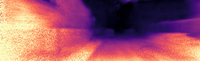} &
		\includegraphics[width=\linewidth]{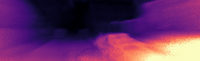} &
		\includegraphics[width=\linewidth]{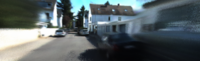} \\
		& \best{SceneRF} &  \includegraphics[width=\linewidth]{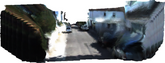} &
		\includegraphics[width=\linewidth]{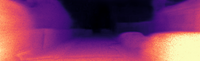} &
		\includegraphics[width=\linewidth]{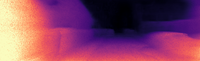} &
		\includegraphics[width=\linewidth]{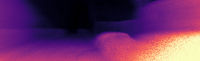} &
		\includegraphics[width=\linewidth]{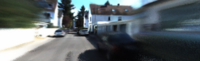} \\
		
		\arrayrulecolor{gray!50}
		\midrule
		
		\multirow{13}{*}{\includegraphics[width=\linewidth]{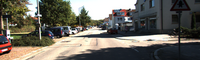}} 
		
		& PixelNeRF & \includegraphics[width=\linewidth]{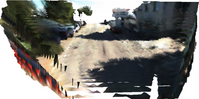} &
		\includegraphics[width=\linewidth]{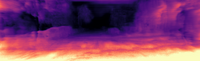} &
		\includegraphics[width=\linewidth]{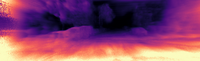} &
		\includegraphics[width=\linewidth]{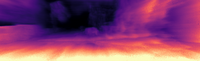} &
		\includegraphics[width=\linewidth]{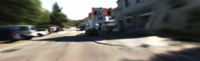} \\
		& MINE & \includegraphics[width=\linewidth]{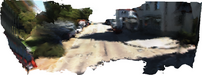} &
		\includegraphics[width=\linewidth]{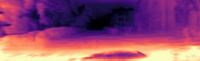} &
		\includegraphics[width=\linewidth]{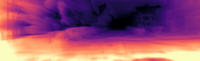} &
		\includegraphics[width=\linewidth]{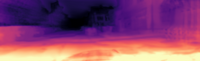} &
		\includegraphics[width=\linewidth]{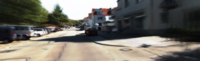} \\
		& VisionNeRF & \includegraphics[width=\linewidth]{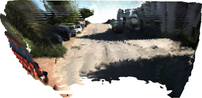} &
		\includegraphics[width=\linewidth]{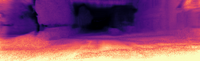} &
		\includegraphics[width=\linewidth]{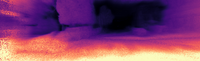} &
		\includegraphics[width=\linewidth]{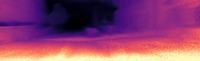} &
		\includegraphics[width=\linewidth]{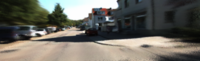} \\
		
		& \best{SceneRF} &  \includegraphics[width=\linewidth]{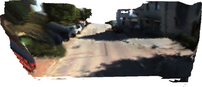} &
		\includegraphics[width=\linewidth]{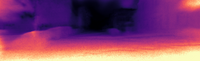} &
		\includegraphics[width=\linewidth]{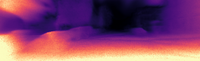} &
		\includegraphics[width=\linewidth]{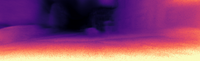} &
		\includegraphics[width=\linewidth]{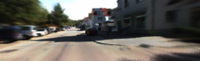} \\
		\arrayrulecolor{gray!50}
		\midrule

	\end{tabular}		
	\caption{\textbf{Additional qualitative results on SemanticKITTI~\cite{semkitti} (val.)}. }
	\label{fig:addtitonal_qualitative_kitti_2}
\end{figure*}

\paragraph{SemanticKITTI~\cite{semkitti}.} We show additional qualitative results in~\cref{fig:addtitonal_qualitative_kitti_1} and~\cref{fig:addtitonal_qualitative_kitti_2}. Overall, our method predicts smoother and finer depth maps, especially at far, which leads to a better-structured 3D scene with fewer artifacts than the baselines.
When synthesizing RGB images, our approach achieves comparable results to other baseline methods.

\begin{figure*}[!t]
	\centering
	\newcolumntype{P}[1]{>{\centering\arraybackslash}m{#1}}
	\setlength{\tabcolsep}{0.004\textwidth}
	\renewcommand{\arraystretch}{0.8}
	\footnotesize
	\begin{tabular}{P{0.12\textwidth}  P{0.10\textwidth}  P{0.18\textwidth}  P{0.14\textwidth} P{0.14\textwidth} P{0.14\textwidth}  P{0.14\textwidth}}	
		\multirow{2}{*}{Input}  & \multirow{2}{*}{Method}  & \multirow{2}{*}{\makecell{3D mesh\\(w/ our recons. scheme)}}  &  \multicolumn{3}{c}{Novel depth}     &  {Novel view} \\
		\cmidrule{4-6}	
		& & & $+0.2$m, $0^{\circ}$ & $+0.2$m, ${-}20^{\circ}$ & ${+}0.4$m, ${+}20^{\circ}$ & ${+}0.4$m, ${+}20^{\circ}$ \\[-0.1em]
		
		\multirow{13}{*}{\includegraphics[width=\linewidth]{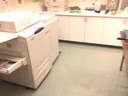}}
		& PixelNeRF & \includegraphics[width=\linewidth]{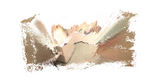} &
		\includegraphics[width=0.65\linewidth]{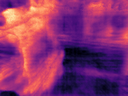} &
		\includegraphics[width=0.65\linewidth]{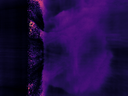} &
		\includegraphics[width=0.65\linewidth]{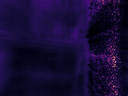} &
		\includegraphics[width=0.65\linewidth]{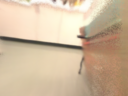} \\
		& MINE & \includegraphics[width=\linewidth]{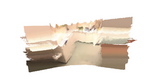} &
		\includegraphics[width=0.65\linewidth]{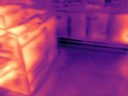} &
		\includegraphics[width=0.65\linewidth]{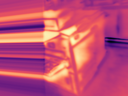} &
		\includegraphics[width=0.65\linewidth]{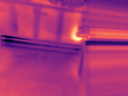} &
		\includegraphics[width=0.65\linewidth]{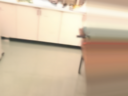} \\
		& VisionNeRF & \includegraphics[width=\linewidth]{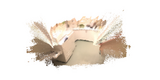} &
		\includegraphics[width=0.65\linewidth]{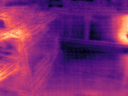} &
		\includegraphics[width=0.65\linewidth]{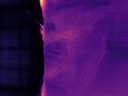} &
		\includegraphics[width=0.65\linewidth]{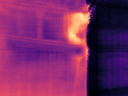} &
		\includegraphics[width=0.65\linewidth]{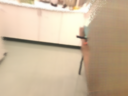} \\
		& \best{SceneRF} &  
		\includegraphics[width=\linewidth]{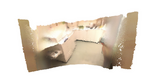} &
		\includegraphics[width=0.65\linewidth]{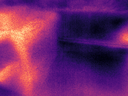} &
		\includegraphics[width=0.65\linewidth]{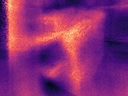} &
		\includegraphics[width=0.65\linewidth]{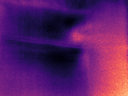} &
		\includegraphics[width=0.65\linewidth]{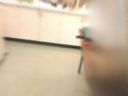} \\
		\arrayrulecolor{gray!50}
		\midrule
		
		\multirow{13}{*}{\includegraphics[width=\linewidth]{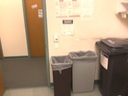}}
		& PixelNeRF & \includegraphics[width=\linewidth]{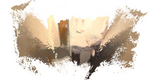} &
		\includegraphics[width=0.65\linewidth]{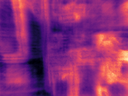} &
		\includegraphics[width=0.65\linewidth]{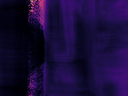} &
		\includegraphics[width=0.65\linewidth]{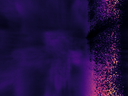} &
		\includegraphics[width=0.65\linewidth]{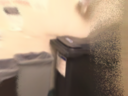} \\
		& MINE & \includegraphics[width=\linewidth]{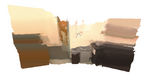} &
		\includegraphics[width=0.65\linewidth]{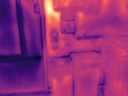} &
		\includegraphics[width=0.65\linewidth]{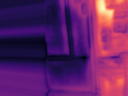} &
		\includegraphics[width=0.65\linewidth]{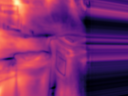} &
		\includegraphics[width=0.65\linewidth]{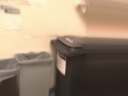} \\
		& VisionNeRF & \includegraphics[width=\linewidth]{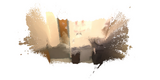} &
		\includegraphics[width=0.65\linewidth]{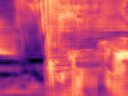} &
		\includegraphics[width=0.65\linewidth]{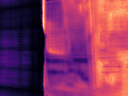} &
		\includegraphics[width=0.65\linewidth]{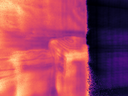} &
		\includegraphics[width=0.65\linewidth]{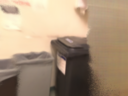} \\
		& \best{SceneRF} &  
		\includegraphics[width=\linewidth]{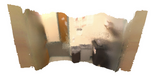} &
		\includegraphics[width=0.65\linewidth]{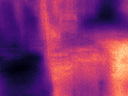} &
		\includegraphics[width=0.65\linewidth]{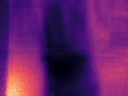} &
		\includegraphics[width=0.65\linewidth]{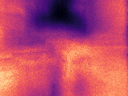} &
		\includegraphics[width=0.65\linewidth]{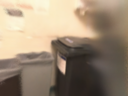} \\
		
		\arrayrulecolor{gray!50}
		\midrule
		
		\multirow{13}{*}{\includegraphics[width=\linewidth]{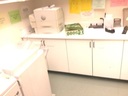}}
		& PixelNeRF & \includegraphics[width=\linewidth]{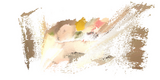} &
		\includegraphics[width=0.65\linewidth]{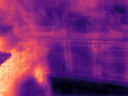} &
		\includegraphics[width=0.65\linewidth]{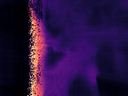} &
		\includegraphics[width=0.65\linewidth]{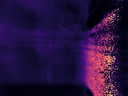} &
		\includegraphics[width=0.65\linewidth]{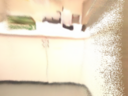} \\
		& MINE & \includegraphics[width=\linewidth]{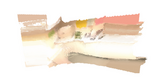} &
		\includegraphics[width=0.65\linewidth]{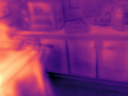} &
		\includegraphics[width=0.65\linewidth]{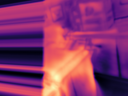} &
		\includegraphics[width=0.65\linewidth]{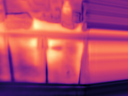} &
		\includegraphics[width=0.65\linewidth]{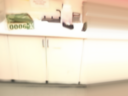} \\
		& VisionNeRF & \includegraphics[width=\linewidth]{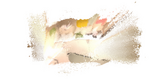} &
		\includegraphics[width=0.65\linewidth]{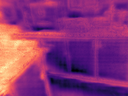} &
		\includegraphics[width=0.65\linewidth]{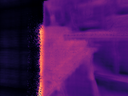} &
		\includegraphics[width=0.65\linewidth]{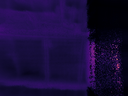} &
		\includegraphics[width=0.65\linewidth]{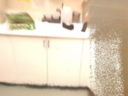} \\
		& \best{SceneRF} &  
		\includegraphics[width=\linewidth]{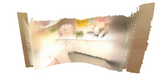} &
		\includegraphics[width=0.65\linewidth]{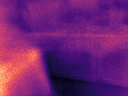} &
		\includegraphics[width=0.65\linewidth]{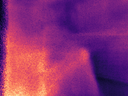} &
		\includegraphics[width=0.65\linewidth]{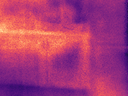} &
		\includegraphics[width=0.65\linewidth]{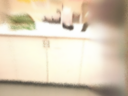} \\

	\end{tabular}		
	\caption{\textbf{Additional qualitative results on BundleFusion~\cite{dai2017bundlefusion} (val.).} }
	\label{fig:supp:qualitative_bundlefusion}
\end{figure*}

\begin{figure*}[!t]
	\centering
	\newcolumntype{P}[1]{>{\centering\arraybackslash}m{#1}}
	\setlength{\tabcolsep}{0.004\textwidth}
	\renewcommand{\arraystretch}{0.8}
	\footnotesize
	\begin{tabular}{P{0.12\textwidth}  P{0.18\textwidth}  P{0.14\textwidth} P{0.14\textwidth} P{0.14\textwidth}  P{0.14\textwidth}}		
		\multirow{2}{*}{Input}   & \multirow{2}{*}{3D mesh}  &  \multicolumn{3}{c}{Novel depth}     &  {Novel view} \\
		\cmidrule{3-5}%
		& &  $+1$m, $0^{\circ}$ & $+3$m, ${-}10^{\circ}$ & ${+}5$m, ${+}10^{\circ}$ & ${+}5$m, ${+}10^{\circ}$ \\

		\includegraphics[width=\linewidth]{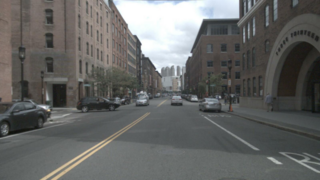} &  
		\includegraphics[width=\linewidth]{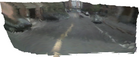} &
		\includegraphics[width=\linewidth]{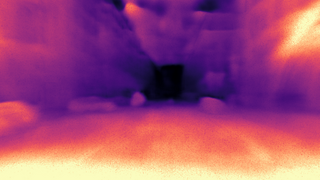} &
		\includegraphics[width=\linewidth]{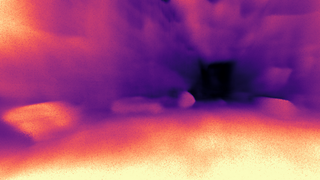} &
		\includegraphics[width=\linewidth]{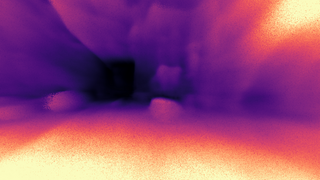} &
		\includegraphics[width=\linewidth]{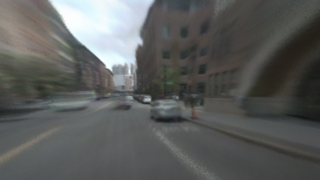} \\

		\includegraphics[width=\linewidth]{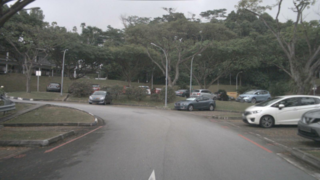} &  
		\includegraphics[width=\linewidth]{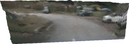} &
		\includegraphics[width=\linewidth]{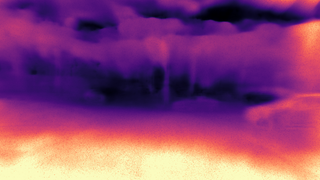} &
		\includegraphics[width=\linewidth]{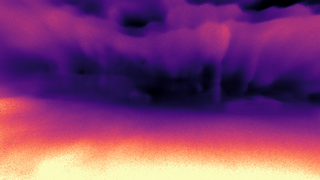} &
		\includegraphics[width=\linewidth]{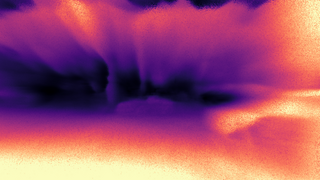} &
		\includegraphics[width=\linewidth]{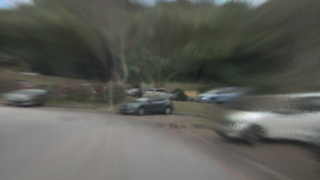} \\

		\includegraphics[width=\linewidth]{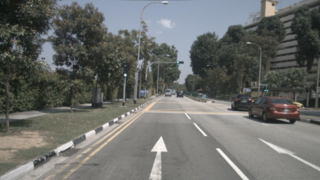} &  
		\includegraphics[width=\linewidth]{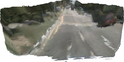} &
		\includegraphics[width=\linewidth]{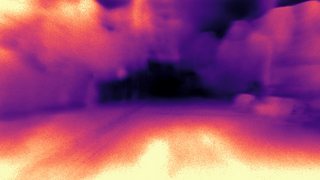} &
		\includegraphics[width=\linewidth]{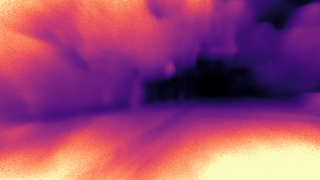} &
		\includegraphics[width=\linewidth]{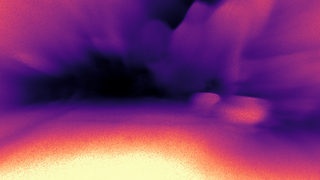} &
		\includegraphics[width=\linewidth]{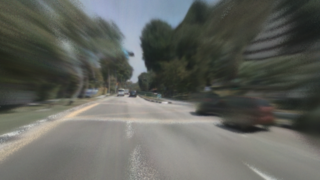} \\

		\includegraphics[width=\linewidth]{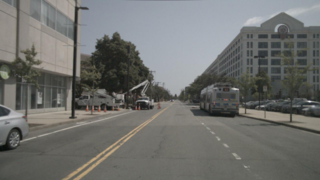} &  
		\includegraphics[width=\linewidth]{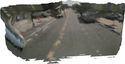} &
		\includegraphics[width=\linewidth]{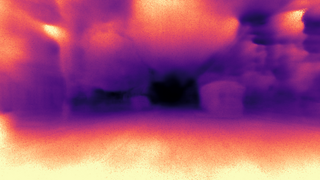} &
		\includegraphics[width=\linewidth]{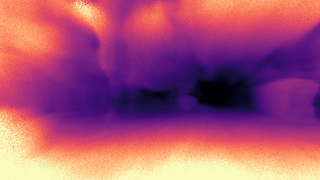} &
		\includegraphics[width=\linewidth]{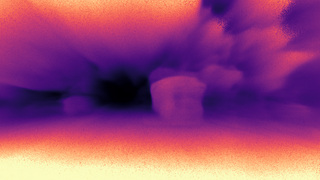} &
		\includegraphics[width=\linewidth]{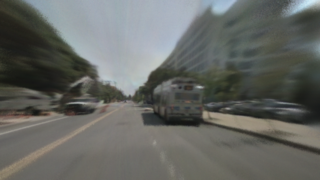} \\

		\includegraphics[width=\linewidth]{figures/supp/nuscenes/inputs/a98fba72bde9433fb882032d18aedb2e.png} &  
		\includegraphics[width=\linewidth]{figures/supp/nuscenes/3d/a98fba72bde9433fb882032d18aedb2e.png} &
		\includegraphics[width=\linewidth]{figures/supp/nuscenes/depths/a98fba72bde9433fb882032d18aedb2e_1.0_0.png} &
		\includegraphics[width=\linewidth]{figures/supp/nuscenes/depths/a98fba72bde9433fb882032d18aedb2e_3.0_-10.png} &
		\includegraphics[width=\linewidth]{figures/supp/nuscenes/depths/a98fba72bde9433fb882032d18aedb2e_5.0_10.png} &
		\includegraphics[width=\linewidth]{figures/supp/nuscenes/rgbs/a98fba72bde9433fb882032d18aedb2e_5.0_10.png} \\

		\includegraphics[width=\linewidth]{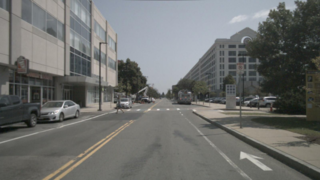} &  
		\includegraphics[width=\linewidth]{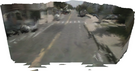} &
		\includegraphics[width=\linewidth]{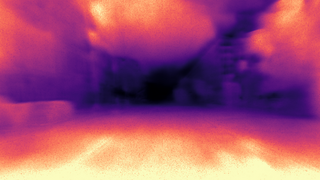} &
		\includegraphics[width=\linewidth]{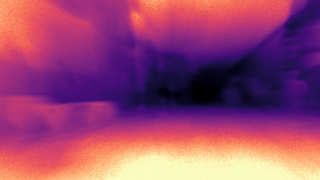} &
		\includegraphics[width=\linewidth]{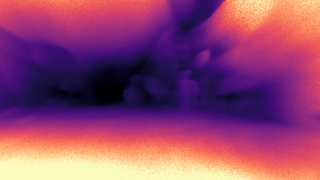} &
		\includegraphics[width=\linewidth]{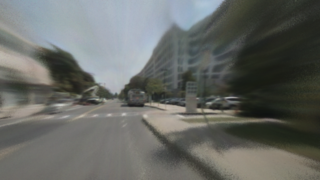} \\

		\includegraphics[width=\linewidth]{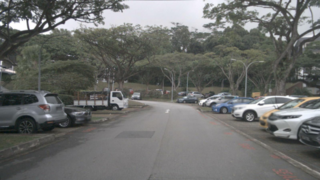} &  
		\includegraphics[width=\linewidth]{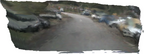} &
		\includegraphics[width=\linewidth]{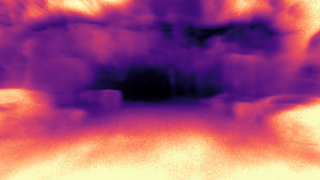} &
		\includegraphics[width=\linewidth]{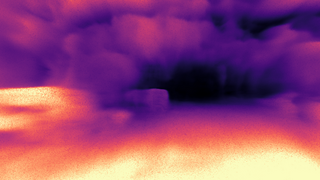} &
		\includegraphics[width=\linewidth]{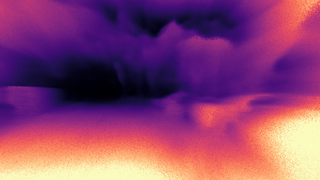} &
		\includegraphics[width=\linewidth]{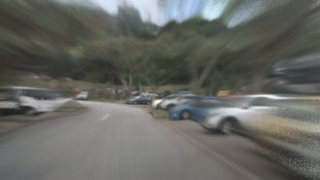} \\

	\end{tabular}		
	\caption{\textbf{Additional generalization results on nuScenes~\cite{Caesar2020nuScenesAM}}. The model is trained only on SemanticKITTI~\cite{semkitti}.}
	\label{fig:supp:nuscenes}
\end{figure*}

\paragraph{BundleFusion~\cite{dai2017bundlefusion}.} We present further qualitative results in~\cref{fig:supp:qualitative_bundlefusion}, which demonstrates that SceneRF produces more accurate depth maps than other methods, especially for views that significantly differ from the input view (as observed in columns ``+0.2m, -20$^{\circ}$'' and ``+0.4m, +20$^{\circ}$''). Moreover, SceneRF is the only method capable of inferring scenery that is not visible in the input field-of-view, which is particularly evident in the first and third rows.
\paragraph{Generalization results on nuScenes~\cite{Caesar2020nuScenesAM}.} Using the SceneRF model trained on Semantic KITTI~\cite{semkitti}, we perform inference on \emph{unseen} nuScenes images. Additional qualitative results obtained are presented in~\cref{fig:supp:nuscenes}. Despite the vastly different setup between the two datasets, such as differences in camera setups and locations (Germany versus USA), SceneRF is able to predict a reasonable scene structure that included \eg, road, building, and vehicles.

{\small
	\bibliographystyle{ieee_fullname}
	\bibliography{egbib}
}

\end{document}

%% file: macros.tex
\newcommand{\ray}{\mathbf{r}}

\newcommand{\Cest}{\hat{C}(\ray)}
\newcommand{\Dest}{\hat{D}(\ray)}

\newcommand{\ngauss}{k}

\definecolor{inputcol}{rgb}{0.24,0.57,0.00}
\definecolor{novelcol}{rgb}{0.65,0.00,0.85}

\newcommand{\xray}{\mathcal{r}_x}
\newcommand{\yray}{\mathcal{r}_y}
\newcommand{\xpix}{x}
\newcommand{\ypix}{y}

\newcommand{\gauss}{\mathcal{G}}

\newcommand{\Lreproj}{\mathcal{L}_{\text{reproj}}}
\newcommand{\Lrgb}{\mathcal{L}_{\text{rgb}}}
\newcommand{\Lsamp}{\mathcal{L}_{\text{samp}}}

\newcommand{\proj}{\texttt{proj}}

\newcommand{\best}[1]{\textbf{#1}}
\newcommand{\second}[1]{\underline{#1}}

%% file: main.bbl
\begin{thebibliography}{10}\itemsep=-1pt

\bibitem{achlioptas2018poincloud}
Panos Achlioptas, Olga Diamanti, Ioannis Mitliagkas, and Leonidas Guibas.
\newblock Learning representations and generative models for 3{D} point clouds.
\newblock In {\em ICML}, 2018.

\bibitem{prsom}
Fatiha Anouar, Fouad Badran, and Sylvie Thiria.
\newblock Probabilistic self-organizing map and radial basis function networks.
\newblock {\em Neurocomputing}, 1998.

\bibitem{barron2022mipnerf360}
Jonathan~T. Barron, Ben Mildenhall, Dor Verbin, Pratul~P. Srinivasan, and Peter
  Hedman.
\newblock Mip-nerf 360: Unbounded anti-aliased neural radiance fields.
\newblock In {\em CVPR}, 2022.

\bibitem{semkitti}
J. Behley, M. Garbade, A. Milioto, J. Quenzel, S. Behnke, C. Stachniss, and J.
  Gall.
\newblock {SemanticKITTI: A Dataset for Semantic Scene Understanding of LiDAR
  Sequences}.
\newblock In {\em ICCV}, 2019.

\bibitem{adabin}
Shariq~Farooq Bhat, Ibraheem Alhashim, and Peter Wonka.
\newblock {AdaBins}: Depth estimation using adaptive bins.
\newblock In {\em CVPR}, 2021.

\bibitem{Caesar2020nuScenesAM}
Holger Caesar, Varun Bankiti, Alex~H. Lang, Sourabh Vora, Venice~Erin Liong,
  Qiang Xu, Anush Krishnan, Yu Pan, Giancarlo Baldan, and Oscar Beijbom.
\newblock nuscenes: A multimodal dataset for autonomous driving.
\newblock In {\em CVPR}, 2020.

\bibitem{Cao2022FWDRN}
Ang Cao, C. Rockwell, and Justin Johnson.
\newblock Fwd: Real-time novel view synthesis with forward warping and depth.
\newblock In {\em CVPR}, 2022.

\bibitem{monoscene}
Anh-Quan Cao and Raoul de Charette.
\newblock Monoscene: Monocular 3d semantic scene completion.
\newblock In {\em CVPR}, 2022.

\bibitem{3DSketch}
Xiaokang Chen, Kwan-Yee Lin, Chen Qian, Gang Zeng, and Hongsheng Li.
\newblock 3d sketch-aware semantic scene completion via semi-supervised
  structure prior.
\newblock In {\em CVPR}, 2020.

\bibitem{chen2020bspnet}
Zhiqin Chen, Andrea Tagliasacchi, and Hao Zhang.
\newblock Bsp-net: Generating compact meshes via binary space partitioning.
\newblock In {\em CVPR}, 2020.

\bibitem{oldtsdffusion}
Brian Curless and Marc Levoy.
\newblock A volumetric method for building complex models from range images.
\newblock In {\em SIGGRAPH}, 1996.

\bibitem{dahnert2021panoptic}
Manuel Dahnert, Ji Hou, Matthias Nießner, and Angela Dai.
\newblock Panoptic 3d scene reconstruction from a single rgb image.
\newblock In {\em NeurIPS}, 2021.

\bibitem{dai2017bundlefusion}
Angela Dai, Matthias Nie{\ss}ner, Michael Zoll{\"o}fer, Shahram Izadi, and
  Christian Theobalt.
\newblock Bundlefusion: Real-time globally consistent 3d reconstruction using
  on-the-fly surface re-integration.
\newblock {\em ACM Transactions on Graphics 2017 (TOG)}, 2017.

\bibitem{dsnerf}
Kangle Deng, Andrew Liu, Jun-Yan Zhu, and Deva Ramanan.
\newblock Depth-supervised {NeRF}: Fewer views and faster training for free.
\newblock In {\em CVPR}, 2022.

\bibitem{Duggal2022TopologicallyAwareDF}
Shivam Duggal and Deepak Pathak.
\newblock Topologically-aware deformation fields for single-view 3d
  reconstruction.
\newblock In {\em CVPR}, 2022.

\bibitem{Ebrahimi2022DifferentiableGS}
Sayna Ebrahimi, Angjoo Kanazawa, and Trevor Darrell.
\newblock Differentiable gradient sampling for learning implicit 3d scene
  reconstructions from a single image.
\newblock In {\em ICLR}, 2022.

\bibitem{Fan_2017_CVPR}
Haoqiang Fan, Hao Su, and Leonidas~J. Guibas.
\newblock A point set generation network for 3d object reconstruction from a
  single image.
\newblock In {\em CVPR}, 2017.

\bibitem{kitticvpr}
Andreas Geiger, Philip Lenz, and Raquel Urtasun.
\newblock Are we ready for autonomous driving? the kitti vision benchmark
  suite.
\newblock In {\em CVPR}, 2012.

\bibitem{gkioxari2020meshRCNN}
Georgia Gkioxari, Jitendra Malik, and Justin Johnson.
\newblock Mesh r-cnn.
\newblock In {\em ICCV}, 2019.

\bibitem{monodepth17}
Cl{\'{e}}ment Godard, Oisin {Mac Aodha}, and Gabriel~J. Brostow.
\newblock Unsupervised monocular depth estimation with left-right consistency.
\newblock In {\em CVPR}, 2017.

\bibitem{monodepth2}
Cl{\'{e}}ment Godard, Oisin {Mac Aodha}, Michael Firman, and Gabriel~J.
  Brostow.
\newblock Digging into self-supervised monocular depth prediction.
\newblock In {\em ICCV}, 2019.

\bibitem{grabner2019location}
Alexander Grabner, Peter~M Roth, and Vincent Lepetit.
\newblock Location field descriptors: Single image 3d model retrieval in the
  wild.
\newblock In {\em 3DV}, 2019.

\bibitem{Han2020DRWRAD}
Zhizhong Han, Chao Chen, Yu-Shen Liu, and Matthias Zwicker.
\newblock Drwr: A differentiable renderer without rendering for unsupervised 3d
  structure learning from silhouette images.
\newblock In {\em ICML}, 2020.

\bibitem{horry1997tour}
Youichi Horry, Ken-Ichi Anjyo, and Kiyoshi Arai.
\newblock Tour into the picture: using a spidery mesh interface to make
  animation from a single image.
\newblock In {\em SIGGRAPH}, 1997.

\bibitem{hu2022efficientnerf}
Tao Hu, Shu Liu, Yilun Chen, Tiancheng Shen, and Jiaya Jia.
\newblock Efficientnerf efficient neural radiance fields.
\newblock In {\em CVPR}, 2022.

\bibitem{huang2018holistic}
Siyuan Huang, Siyuan Qi, Yixin Zhu, Yinxue Xiao, Yuanlu Xu, and Song-Chun Zhu.
\newblock Holistic 3d scene parsing and reconstruction from a single rgb image.
\newblock In {\em ECCV}, 2018.

\bibitem{tpvformer}
Yuanhui Huang, Wenzhao Zheng, Yunpeng Zhang, Jie Zhou, and Jiwen Lu.
\newblock Tri-perspective view for vision-based 3d semantic occupancy
  prediction.
\newblock In {\em CVPR}, 2023.

\bibitem{Huang2022PlanesVC}
Zixuan Huang, Stefan Stojanov, Anh Thai, Varun Jampani, and James~M. Rehg.
\newblock Planes vs. chairs: Category-guided 3d shape learning without any 3d
  cues.
\newblock In {\em ECCV}, 2022.

\bibitem{izadinia2017im2cad}
Hamid Izadinia, Qi Shan, and Steven~M. Seitz.
\newblock Im2cad.
\newblock In {\em CVPR}, 2017.

\bibitem{codenerf}
Wonbong Jang and Lourdes Agapito.
\newblock Codenerf: Disentangled neural radiance fields for object categories.
\newblock In {\em ICCV}, 2021.

\bibitem{koenderink1995depth}
Jan~J Koenderink, Andrea~J van Doorn, and Astrid~ML Kappers.
\newblock Depth relief.
\newblock {\em Perception}, 1995.

\bibitem{Kurz2022AdaNeRFAS}
Andreas Kurz, Thomas Neff, Zhaoyang Lv, Michael Zollhofer, and Markus
  Steinberger.
\newblock Adanerf: Adaptive sampling for real-time rendering of neural radiance
  fields.
\newblock In {\em ECCV}, 2022.

\bibitem{lee2017roomnet}
Chen-Yu Lee, Vijay Badrinarayanan, Tomasz Malisiewicz, and Andrew Rabinovich.
\newblock Roomnet: End-to-end room layout estimation.
\newblock In {\em ICCV}, 2017.

\bibitem{mine2021}
Jiaxin Li, Zijian Feng, Qi She, Henghui Ding, Changhu Wang, and Gim~Hee Lee.
\newblock Mine: Towards continuous depth mpi with nerf for novel view
  synthesis.
\newblock In {\em ICCV}, 2021.

\bibitem{aicnet}
Jie Li, Kai Han, Peng Wang, Yu Liu, and Xia Yuan.
\newblock Anisotropic convolutional networks for 3d semantic scene completion.
\newblock In {\em CVPR}, 2020.

\bibitem{symmnerf}
Xingyi Li, Chaoyi Hong, Yiran Wang, Zhiguo Cao, Ke Xian, and Guosheng Lin.
\newblock Symmnerf: Learning to explore symmetry prior for single-view view
  synthesis.
\newblock In {\em ACCV}, 2022.

\bibitem{li2023voxformer}
Yiming Li, Zhiding Yu, Christopher Choy, Chaowei Xiao, Jose~M Alvarez, Sanja
  Fidler, Chen Feng, and Anima Anandkumar.
\newblock Voxformer: Sparse voxel transformer for camera-based 3d semantic
  scene completion.
\newblock In {\em CVPR}, 2023.

\bibitem{deepmarchingcube}
Yiyi Liao, Simon Donné, and Andreas Geiger.
\newblock Deep marching cubes: Learning explicit surface representations.
\newblock In {\em CVPR}, 2018.

\bibitem{visionnerf}
Kai-En Lin, Yen-Chen Lin, Wei-Sheng Lai, Tsung-Yi Lin, Yichang Shih, and Ravi
  Ramamoorthi.
\newblock Vision transformer for nerf-based view synthesis from a single input
  image.
\newblock In {\em WACV}, 2023.

\bibitem{Liu20222DGM}
Feng Liu and Xiaoming Liu.
\newblock 2d gans meet unsupervised single-view 3d reconstruction.
\newblock In {\em ECCV}, 2022.

\bibitem{marchingcubes}
William~E Lorensen and Harvey~E Cline.
\newblock Marching cubes: A high resolution 3d surface construction algorithm.
\newblock In {\em SIGGRAPH}, 1987.

\bibitem{adamW}
Ilya Loshchilov and Frank Hutter.
\newblock Decoupled weight decay regularization.
\newblock In {\em ICLR}, 2019.

\bibitem{miao2023occdepth}
Ruihang Miao, Weizhou Liu, Mingrui Chen, Zheng Gong, Weixin Xu, Chen Hu, and
  Shuchang Zhou.
\newblock Occdepth: A depth-aware method for 3d semantic scene completion.
\newblock In {\em CVPR}, 2023.

\bibitem{mildenhall2019llff}
Ben Mildenhall, Pratul~P. Srinivasan, Rodrigo Ortiz-Cayon, Nima~Khademi
  Kalantari, Ravi Ramamoorthi, Ren Ng, and Abhishek Kar.
\newblock Local light field fusion: Practical view synthesis with prescriptive
  sampling guidelines.
\newblock {\em TOG}, 2019.

\bibitem{nerf}
Ben Mildenhall, Pratul~P. Srinivasan, Matthew Tancik, Jonathan~T. Barron, Ravi
  Ramamoorthi, and Ren Ng.
\newblock Nerf: Representing scenes as neural radiance fields for view
  synthesis.
\newblock In {\em ECCV}, 2020.

\bibitem{ming2021deep}
Yue Ming, Xuyang Meng, Chunxiao Fan, and Hui Yu.
\newblock Deep learning for monocular depth estimation: A review.
\newblock {\em Neurocomputing}, 2021.

\bibitem{Mller2022AutoRFL3}
Norman M{\"u}ller, Andrea Simonelli, Lorenzo Porzi, Samuel~Rota Bul{\`o},
  Matthias Nie{\ss}ner, and Peter Kontschieder.
\newblock Autorf: Learning 3d object radiance fields from single view
  observations.
\newblock In {\em CVPR}, 2022.

\bibitem{donerf}
Thomas Neff, Pascal Stadlbauer, Mathias Parger, Andreas Kurz, Joerg~H. Mueller,
  Chakravarty R.~Alla Chaitanya, Anton~S. Kaplanyan, and Markus Steinberger.
\newblock {DONeRF: Towards Real-Time Rendering of Compact Neural Radiance
  Fields using Depth Oracle Networks}.
\newblock {\em CGF}, 2021.

\bibitem{kinectfusion}
Richard~A. Newcombe, Shahram Izadi, Otmar Hilliges, David Molyneaux, David Kim,
  Andrew~J. Davison, Pushmeet Kohli, Jamie Shotton, Steve Hodges, and Andrew~W.
  Fitzgibbon.
\newblock Kinectfusion: Real-time dense surface mapping and tracking.
\newblock {\em ISMAR}, 2011.

\bibitem{nie2020total3dunderstanding}
Yinyu Nie, Xiaoguang Han, Shihui Guo, Yujian Zheng, Jian Chang, and Jian~Jun
  Zhang.
\newblock Total3dunderstanding: Joint layout, object pose and mesh
  reconstruction for indoor scenes from a single image.
\newblock In {\em CVPR}, 2020.

\bibitem{Niemeyer2020DifferentiableVR}
Michael Niemeyer, Lars~M. Mescheder, Michael Oechsle, and Andreas Geiger.
\newblock Differentiable volumetric rendering: Learning implicit 3d
  representations without 3d supervision.
\newblock In {\em CVPR}, 2020.

\bibitem{kenburn3d}
Simon Niklaus, Long Mai, Jimei Yang, and Feng Liu.
\newblock {3D} ken burns effect from a single image.
\newblock {\em TOG}, 2019.

\bibitem{Park2019DeepSDFLC}
Jeong~Joon Park, Peter Florence, Julian Straub, Richard~A. Newcombe, and Steven
  Lovegrove.
\newblock {DeepSDF}: Learning continuous signed distance functions for shape
  representation.
\newblock In {\em CVPR}, 2019.

\bibitem{PengNMP020convoccnet}
Songyou Peng, Michael Niemeyer, Lars~M. Mescheder, Marc Pollefeys, and Andreas
  Geiger.
\newblock Convolutional occupancy networks.
\newblock In {\em ECCV}, 2020.

\bibitem{Popov2020CoReNetC3}
S. Popov, Pablo Bauszat, and V. Ferrari.
\newblock Corenet: Coherent 3d scene reconstruction from a single rgb image.
\newblock In {\em ECCV}, 2020.

\bibitem{reizenstein2021common}
Jeremy Reizenstein, Roman Shapovalov, Philipp Henzler, Luca Sbordone, Patrick
  Labatut, and David Novotny.
\newblock Common objects in {3D}: Large-scale learning and evaluation of
  real-life {3D} category reconstruction.
\newblock In {\em ICCV}, 2021.

\bibitem{urbannerf}
Konstantinos Rematas, Andrew Liu, Pratul~P. Srinivasan, Jonathan~T. Barron,
  Andrea Tagliasacchi, Tom Funkhouser, and Vittorio Ferrari.
\newblock Urban radiance fields.
\newblock In {\em CVPR}, 2022.

\bibitem{sharf}
Konstantinos Rematas, Ricardo Martin-Brualla, and Vittorio Ferrari.
\newblock Sharf: Shape-conditioned radiance fields from a single view.
\newblock In {\em ICML}, 2021.

\bibitem{roessle2022depthpriorsnerf}
Barbara Roessle, Jonathan~T. Barron, Ben Mildenhall, Pratul~P. Srinivasan, and
  Matthias Nie{\ss}ner.
\newblock Dense depth priors for neural radiance fields from sparse input
  views.
\newblock In {\em CVPR}, 2022.

\bibitem{lmscnet}
Luis Rold{\~a}o, Raoul de Charette, and Anne Verroust-Blondet.
\newblock Lmscnet: Lightweight multiscale 3d semantic completion.
\newblock In {\em 3DV}, 2020.

\bibitem{roldao20223d}
Luis Roldao, Raoul De~Charette, and Anne Verroust-Blondet.
\newblock 3d semantic scene completion: a survey.
\newblock {\em IJCV}, 2022.

\bibitem{salomon2006transformations}
David Salomon.
\newblock {\em Transformations and projections in computer graphics}.
\newblock Springer, 2006.

\bibitem{sharma2022seeing}
Prafull Sharma, Ayush Tewari, Yilun Du, Sergey Zakharov, Rares Ambrus, Adrien
  Gaidon, William~T Freeman, Fredo Durand, Joshua~B Tenenbaum, and Vincent
  Sitzmann.
\newblock Seeing 3d objects in a single image via self-supervised
  static-dynamic disentanglement.
\newblock {\em arXiv preprint arXiv:2207.11232}, 2022.

\bibitem{sitzmann2021lfns}
Vincent Sitzmann, Semon Rezchikov, William~T. Freeman, Joshua~B. Tenenbaum, and
  Fredo Durand.
\newblock Light field networks: Neural scene representations with
  single-evaluation rendering.
\newblock In {\em NeurIPS}, 2021.

\bibitem{sitzmann2019srns}
Vincent Sitzmann, Michael Zollh{\"o}fer, and Gordon Wetzstein.
\newblock Scene representation networks: Continuous 3d-structure-aware neural
  scene representations.
\newblock In {\em NeurIPS}, 2019.

\bibitem{sturm1999method}
Peter Sturm and Steve Maybank.
\newblock A method for interactive 3d reconstruction of piecewise planar
  objects from single images.
\newblock In {\em BMVC}, 1999.

\bibitem{sun2019horizonnet}
Cheng Sun, Chi-Wei Hsiao, Min Sun, and Hwann-Tzong Chen.
\newblock Horizonnet: Learning room layout with 1d representation and pano
  stretch data augmentation.
\newblock In {\em CVPR}, 2019.

\bibitem{lama}
Roman Suvorov, Elizaveta Logacheva, Anton Mashikhin, Anastasia Remizova,
  Arsenii Ashukha, Aleksei Silvestrov, Naejin Kong, Harshith Goka, Kiwoong
  Park, and Victor~S. Lempitsky.
\newblock Resolution-robust large mask inpainting with fourier convolutions.
\newblock In {\em WACV}, 2021.

\bibitem{effcientnet}
Mingxing Tan and Quoc Le.
\newblock {E}fficient{N}et: Rethinking model scaling for convolutional neural
  networks.
\newblock In {\em ICML}, 2019.

\bibitem{tatarchenko2015single}
Maxim Tatarchenko, Alexey Dosovitskiy, and Thomas Brox.
\newblock Single-view to multi-view: Reconstructing unseen views with a
  convolutional network.
\newblock {\em CoRR}, 2015.

\bibitem{grf2021}
Alex Trevithick and Bo Yang.
\newblock {GRF}: Learning a general radiance field for {3D} scene
  representation and rendering.
\newblock In {\em ICCV}, 2021.

\bibitem{VANDENHEUVEL1998354}
Frank~A Van~den Heuvel.
\newblock 3d reconstruction from a single image using geometric constraints.
\newblock {\em ISPRS}, 1998.

\bibitem{wang2018pixel2mesh}
Nanyang Wang, Yinda Zhang, Zhuwen Li, Yanwei Fu, Wei Liu, and Yu-Gang Jiang.
\newblock Pixel2mesh: Generating 3d mesh models from single rgb images.
\newblock In {\em ECCV}, 2018.

\bibitem{ssim}
Zhou Wang, A.C. Bovik, H.R. Sheikh, and E.P. Simoncelli.
\newblock Image quality assessment: from error visibility to structural
  similarity.
\newblock {\em TIP}, 2004.

\bibitem{synsin}
Olivia Wiles, Georgia Gkioxari, Richard Szeliski, and Justin Johnson.
\newblock {SynSin}: {E}nd-to-end view synthesis from a single image.
\newblock In {\em CVPR}, 2020.

\bibitem{wimbauer2023behind}
Felix Wimbauer, Nan Yang, Christian Rupprecht, and Daniel Cremers.
\newblock Behind the scenes: Density fields for single view reconstruction.
\newblock In {\em CVPR}, 2023.

\bibitem{pix2voxplus}
Haozhe Xie, Hongxun Yao, Shengping Zhang, Shangchen Zhou, and Wenxiu Sun.
\newblock Pix2vox++: Multi-scale context-aware 3d object reconstruction from
  single and multiple images.
\newblock {\em IJCV}, 2020.

\bibitem{xie2022neural}
Yiheng Xie, Towaki Takikawa, Shunsuke Saito, Or Litany, Shiqin Yan, Numair
  Khan, Federico Tombari, James Tompkin, Vincent Sitzmann, and Srinath Sridhar.
\newblock Neural fields in visual computing and beyond.
\newblock In {\em EUROGRAPHICS}, 2022.

\bibitem{xu_disn_2019}
Qiangeng Xu, Weiyue Wang, Duygu Ceylan, Radomir Mech, and Ulrich Neumann.
\newblock Disn: Deep implicit surface network for high-quality single-view 3d
  reconstruction.
\newblock In {\em NeurIPS}, 2019.

\bibitem{yang2019pointflow}
Guandao Yang, Xun Huang, Zekun Hao, Ming-Yu Liu, Serge Belongie, and Bharath
  Hariharan.
\newblock Pointflow: 3d point cloud generation with continuous normalizing
  flows.
\newblock In {\em ICCV}, 2019.

\bibitem{pixelnerf}
Alex Yu, Vickie Ye, Matthew Tancik, and Angjoo Kanazawa.
\newblock {pixelNeRF}: Neural radiance fields from one or few images.
\newblock In {\em CVPR}, 2021.

\bibitem{Zakharov2021SingleShotSR}
Sergey Zakharov, Rares Ambrus, Vitor~Campagholo Guizilini, Dennis Park, Wadim
  Kehl, Fr{\'e}do Durand, Joshua~B. Tenenbaum, Vincent Sitzmann, Jiajun Wu, and
  Adrien Gaidon.
\newblock Single-shot scene reconstruction.
\newblock In {\em CoRL}, 2021.

\bibitem{3dmatch}
Andy Zeng, Shuran Song, Matthias Nießner, Matthew Fisher, Jianxiong Xiao, and
  Thomas Funkhouser.
\newblock 3dmatch: Learning local geometric descriptors from rgb-d
  reconstructions.
\newblock In {\em CVPR}, 2017.

\bibitem{790426}
M. Zerroug and R. Nevatia.
\newblock Part-based 3d descriptions of complex objects from a single image.
\newblock {\em TPAMI}, 1999.

\bibitem{zhang2021holistic}
Cheng Zhang, Zhaopeng Cui, Yinda Zhang, Bing Zeng, Marc Pollefeys, and
  Shuaicheng Liu.
\newblock Holistic 3d scene understanding from a single image with implicit
  representation.
\newblock In {\em CVPR}, 2021.

\bibitem{zhang2020perceiving}
Jason~Y Zhang, Sam Pepose, Hanbyul Joo, Deva Ramanan, Jitendra Malik, and
  Angjoo Kanazawa.
\newblock Perceiving 3d human-object spatial arrangements from a single image
  in the wild.
\newblock In {\em ECCV}, 2020.

\bibitem{lpips}
Richard Zhang, Phillip Isola, Alexei~A. Efros, Eli Shechtman, and Oliver Wang.
\newblock The unreasonable effectiveness of deep features as a perceptual
  metric.
\newblock In {\em CVPR}, 2018.

\bibitem{zou2018layoutnet}
Chuhang Zou, Alex Colburn, Qi Shan, and Derek Hoiem.
\newblock Layoutnet: Reconstructing the 3d room layout from a single rgb image.
\newblock In {\em CVPR}, 2018.

\bibitem{Zubic2021AnEL}
Nikola Zubi'c and Pietro Lio’.
\newblock An effective loss function for generating 3d models from single 2d
  image without rendering.
\newblock In {\em AIAI}, 2021.

\end{thebibliography}
